\documentclass[twocolumn]{wlscirep}
\usepackage[utf8]{inputenc}
\usepackage[T1]{fontenc}
\usepackage{ulem}

\usepackage{listings,xcolor}
\usepackage{amsmath,amssymb}
\usepackage{changepage}

\usepackage{multirow}
\usepackage{float}

\title{AGAR a microbial colony dataset for deep learning detection}

\author[1,2]{Sylwia Majchrowska}
\author[1,2]{Jarosław Pawłowski}
\author[1,3]{Grzegorz Guła}
\author[1]{Tomasz Bonus}
\author[1]{Agata Hanas}
\author[1]{Adam Loch}
\author[1]{Agnieszka Pawlak}
\author[3]{Justyna Roszkowiak}
\author[1,*]{Tomasz Golan}
\author[1,3,+]{Zuzanna Drulis-Kawa}
\affil[1]{NeuroSYS, Rybacka 7, 53-656 Wrocław, Poland}
\affil[2]{Faculty of Fundamental Problems of Technology, Wroclaw University of Science and Technology,\newline Wybrzeże S Wyspiańskiego 27, 50-372 Wroclaw, Poland}
\affil[3]{Department of Pathogen Biology and Immunology, Institute of Genetics and Microbiology, University of Wroclaw, \newline S. Przybyszewskiego 63, 51-148 Wroclaw, Poland}

\affil[*]{t.golan@neurosys.com}
\affil[+]{zuzanna.drulis-kawa@uwr.edu.pl}

\begin{abstract}
The Annotated Germs for Automated Recognition (AGAR) dataset is an image database of microbial colonies cultured on agar plates. It contains 18~000 photos of five different microorganisms as single or mixed cultures, taken under diverse lighting conditions with two different cameras. All the images are classified into \textit{countable}, \textit{uncountable}, and \textit{empty}, with the \textit{countable} class labeled by microbiologists with colony location and species identification (336~442 colonies in total). This study describes the dataset itself and the process of its development. In the second part, the performance of selected deep neural network architectures for object detection, namely Faster R-CNN and Cascade R-CNN, was evaluated on the AGAR dataset. The results confirmed the great potential of deep learning methods to automate the process of microbe localization and classification based on Petri dish photos. Moreover, AGAR is the first publicly available dataset of this kind and size and will facilitate the future development of machine learning models. The data used in these studies can be found at \url{https://agar.neurosys.com/}.
\end{abstract}
\begin{document}

\flushbottom
\maketitle
%
%
\thispagestyle{empty}


\section{Introduction}

In recent years, data-driven artificial intelligence (AI) methods have dominated automated pattern search. In particular, deep learning (DL) approaches successfully reduce the need for feature engineering by leveraging a large amount of data to optimize the model’s parameters and find the most important data features~\cite{lecun2015deep}. DL is more and more broadly used in many scientific disciplines. Deep neural networks trained on huge datasets are irreplaceable in modern imaging diagnostics in medicine and allow for fast and precise clinical decisions and medical procedure implementations~\cite{NMI1, NMI3, NMI4, NMI5, NMI6}. Another example is the application of DL in cancer research and diagnostics~\cite{CR1, CR2, CR3, CR4}, where it improves automation and opens new avenues for artificial intelligence-assisted precision oncology and medicine in general. Overall, the implementation of AI provides a new quality of medical tools and diagnostics enabling faster and more efficient healthcare procedures.

Another area where AI could be implemented is microbiology with regard to standard procedural requirements in the pharmaceutical, cosmetic, or food industries. The manufacturing processes are subjected to strict policies and regulations listed, for instance, in Pharmacopoeia~\cite{bib:M1, bib:M2, bib:M3} or European Medicines Agency (EMA) guidelines~\cite{bib:M4}, including extensive regulations governing microbiological purity in industrial areas~\cite{bib:M5, bib:M6}. The above requirements obligate manufacturers to perform constant microbiological monitoring, which means thousands of samples analysed by experienced microbiologists. In most production plants, microbiological procedures are entirely manual - starting from sample collection and culturing to the final plate evaluation with microbial colony count and identification. Large companies start implementing laboratory automation systems to speed up the operating process~\cite{bib:Ferrari2017, bib:FaronCOP}. Taking into account the current industry demand, we aim to develop an effective automatic analysis of microbiological samples based on DL algorithms, performing an efficient and precise microbial sample analysis.

The existing solutions for microbial colony analysis provide only some level of automation. Commonly, the process is limited to generating the photo of a microbial culture grown on a Petri dish, which is further investigated and manually curated by microbiologists. However, the full automation of microbial sample analysis might be accomplished by the implementation of deep neural networks providing robust and accurate algorithms for microbial colony localization, classification, and counting. DL-based solutions can generalize different acquisition setups and new microbe species easier than taditional computer vision algorithms, which strongly depend on hand-crafted feature selection~\cite{ZD1}. As DL requires well-balanced and diverse data to optimally fit the model’s parameters, there is a need to create huge publicly available datasets for neural network training and, consequently, achieving their best performance.

\begin{figure*}[!ht]
\centering
\includegraphics[width=.95\textwidth]{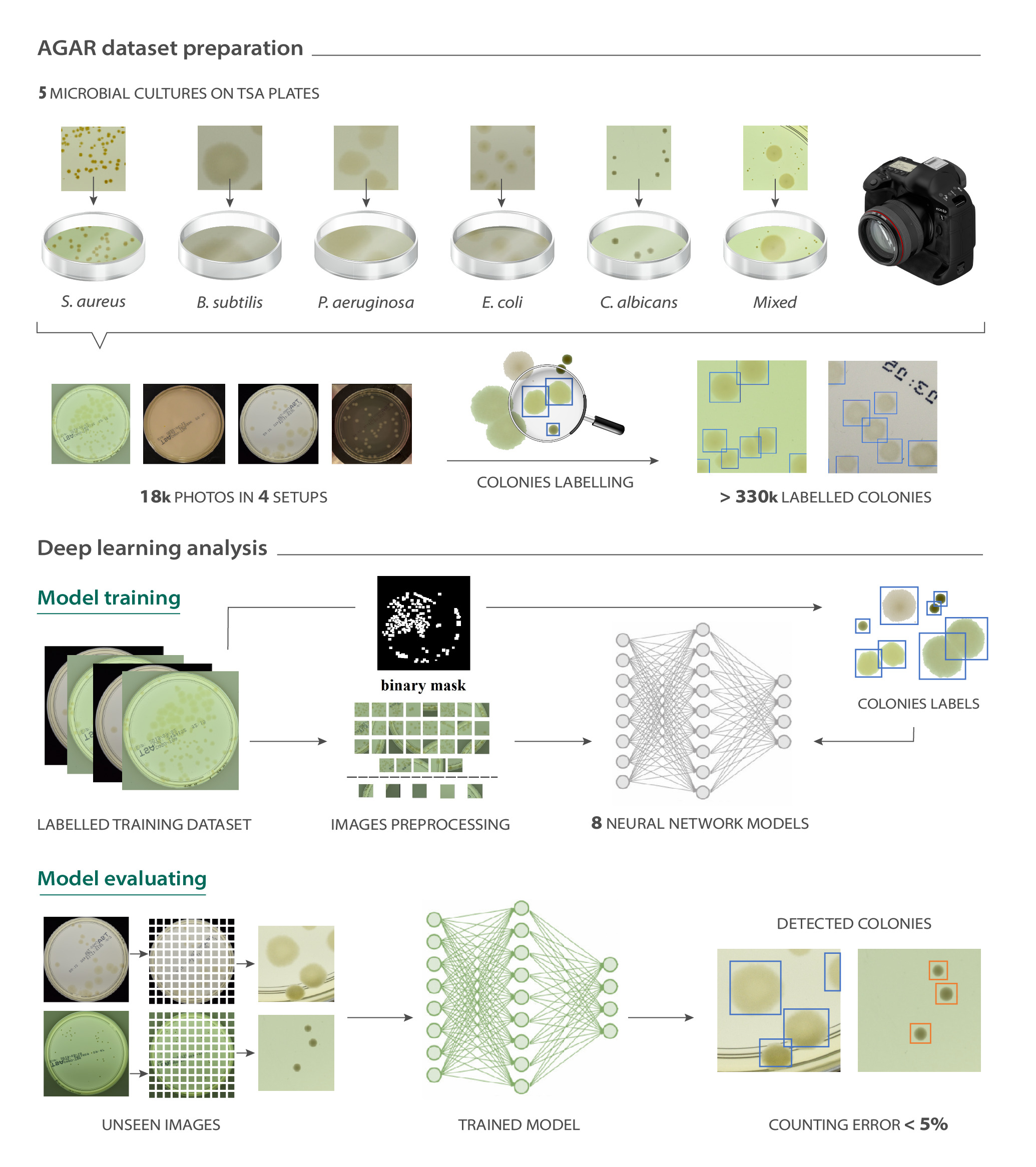}
\caption{\label{fig:graphical_abstract}\textbf{An overview of the processing pipeline from microbiological sample preparation to model performance evaluation.} The \textit{AGAR dataset preparation} (top panel) includes 5 microbial species as single and mixed cultures on agar plates from different liquid dilutions, followed by image acquisition with 4 different setups, finalized with colony labeling and identification by a microbiologist. The \textit{deep learning analysis} (bottom panel) using 8 neural network architectures consists of two separate stages. Images randomly cut into patches constituted the network input in the \textit{model training} stage, while overlapping patches of unseen images were used in the \textit{model evaluating} stage. See technical details in Supplementary Section~2.1.}
\end{figure*}

The main objective of our work was to develop a DL-based methodology to identify and count microbial colonies based on the photo of a standard agar plate culture. At first, we prepared the Annotated Germs for Automated Recognition (AGAR) dataset consisting of over 330k labeled microbial colonies (\textit{Staphylococcus aureus, Bacillus subtilis, Pseudomonas aeruginosa, Escherichia coli}, and \textit{Candida albicans}) localized on 18k annotated Petri dish photos. Secondly, we evaluated the selected deep neural networks, such as Faster R-CNN~\cite{bib:Faster2015} and Cascade R-CNN~\cite{bib:Cascade2018} architectures for object detection, with several different backbones -- ResNet~\cite{bib:resnet2016}, ResNeXt~\cite{bib:resnext2017}, and HRNet~\cite{bib:hrnet2020} -- to set benchmarks for the task. The whole process of collecting and annotating the data as well as the deep learning analysis is presented in Fig.\ref{fig:graphical_abstract}.

\section*{AGAR dataset preparation}

Designing a dataset requires several decisions to be made. In the case of data imaging  for microbial colony counting and identification, one needs to select microorganism species to study, provide proper growth conditions (applied medium, incubation temperature and time, culture dilution factors, etc.), and choose an appropriate photo acquisition setup (light source location, external light handling, or camera type). Such decisions for the AGAR dataset were based on the goal of making a diverse dataset for automatic colony counting on agar plates which could be easily reproduced in any microbiological laboratory.

The selection of standard microorganisms for the AGAR dataset preparation was based on Pharmacopoeia~\cite{bib:M1, bib:M2, bib:M3} regulations and ATCC guidelines~\cite{bib:M15}. Five representatives from different bacterial groups (\textit{S. aureus} subsp. \textit{aureus} ATCC 6538, \textit{B. subtilis} subsp. \textit{spizizenii} ATCC 6633, \textit{P. aeruginosa} ATCC 9027, \textit{E. coli} ATCC 8739), and \textit{C. albicans} ATCC 10231 as a yeast strain were chosen. A series of 10-fold dilutions of refreshed cultures were made and 100 ul were inoculated onto Trypticase Soy Agar (TSA) plates in five technical replicates. The microbial cultures were then incubated at 37\textsuperscript{o}C for 18-24 hours. For data acquisition, single and mixed cultures (\textit{S. aureus} \& \textit{P. aeruginosa/E. coli/C. albicans}; \textit{P. aeruginosa} \& \textit{E. coli/C. albicans}; \textit{E. coli} \& \textit{C. albicans}) were used. The images of colonies grown on agar plates were taken using one of two cameras. Thus, the whole photo dataset was divided into two major subsets: \textit{higher-resolution} (4000 x 6000 px) and \textit{lower-resolution} (2048 x 2048 px). Moreover, three different subgroups were distinguished in the former subset: \textit{bright, dark}, and \textit{vague} based on different illumination conditions in which the photos were taken. For visualization of plates captured in the various setups see Fig.~\ref{fig:boxed}, whereas details on microbial colony growth conditions and acquisition setups are provided in Supplementary Sections 1.1 and 1.2.

\begin{figure*}[!tb]
\centering
\includegraphics[width=\textwidth]{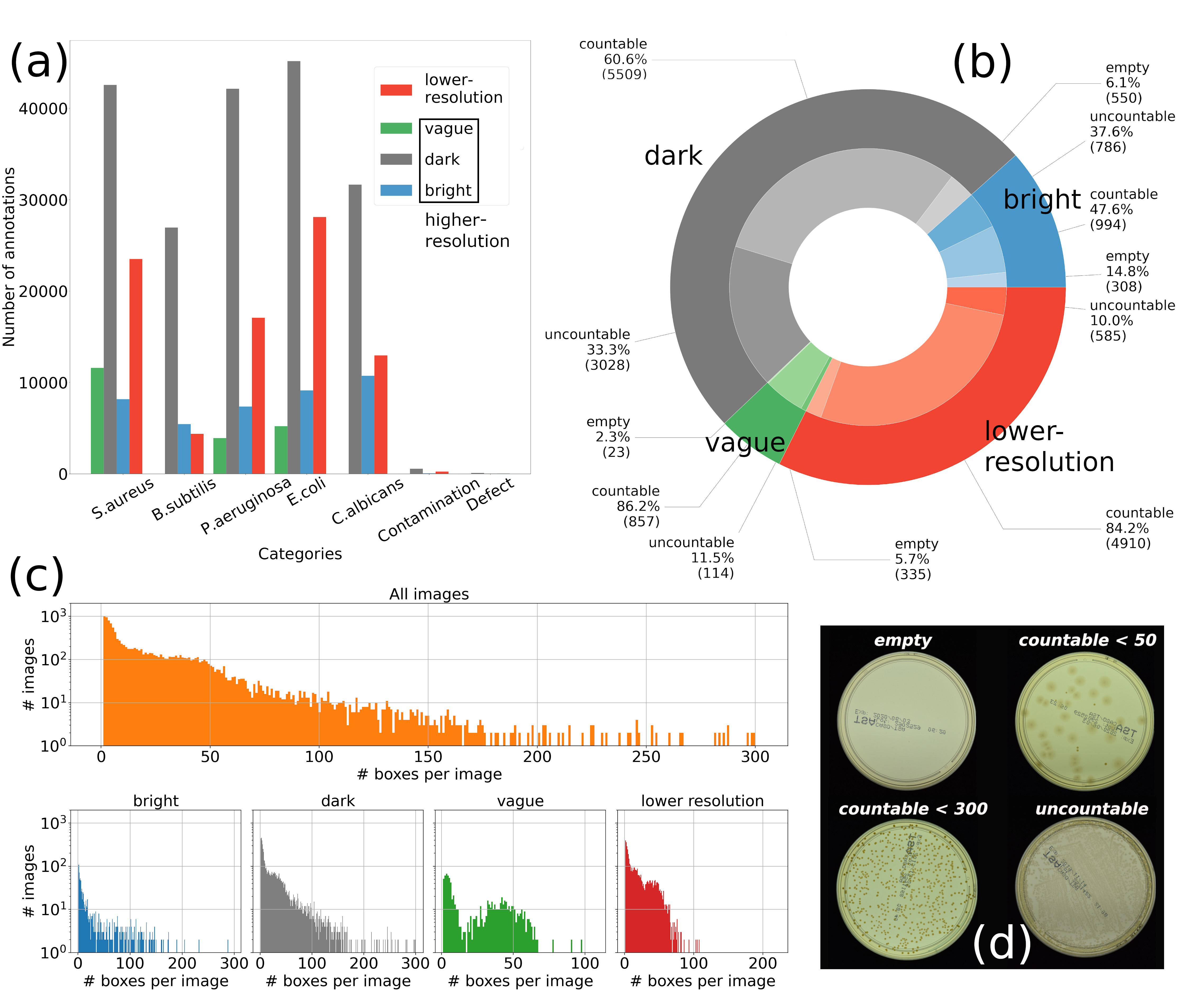}
\caption{\label{fig:statistics}\textbf{The AGAR dataset statistics summary.} The number of samples distribution: (a) over different microbial species, and (b) for different acquisition setup subgroups. The combination of samples from \textit{bright} and \textit{dark} subgroups constitutes about half of the whole dataset and includes photos with the best quality in terms of the contrast between grown microorganisms and agar surface. (c) The histograms of the number of annotated instances per image calculated including \textit{countable} samples for different subsets separately (blue, grey, green, red) and altogether (orange). The mean number of annotations per image equals 27.4. (d) Image examples of agar plates with different numbers of growing colonies, marked as \textit{empty, countable}, and \textit{uncountable}.}
\end{figure*}

A web application was developed for microbiologists to upload and annotate agar plate culture photos. Each sample recorded in the database was marked as \textit{countable, uncountable}, or \textit{empty} (Fig.~\ref{fig:statistics}(d)). It was important to establish a countable range for manual counting, which is generally between 30 - 300 colonies for a standard 100 mm Petri dish~\cite{bib:Tomasiewicz1980, bib:Sutton2011}. Every \textit{countable} sample was manually labeled by microbiologists using a bounding box and species label per colony. Samples were classified as \textit{uncountable} if more than 300 colonies were present, or if no discrete colonies were seen due to their agglomeration or blurred shape.

In total, 18 000 photos taken under diverse lighting conditions were collected in the AGAR database. Altogether, 336~442 colonies of 5 standard microbes were annotated. The summary of the AGAR dataset statistics is presented in Fig.~\ref{fig:statistics}(b). The AGAR dataset achieves a good balance in the number of instances for different microbes (see Fig.~\ref{fig:statistics}(a)), which is important for building a robust deep learning model. The interquartile range for all 12~271 countable samples is between 4 and 38 colonies, and around 84.4\% of these samples have less than 50 colonies. The histograms of the number of annotated colonies per \textit{countable} image are presented in Fig.~\ref{fig:statistics}(c). Additional information of provided annotations and bounding box size distribution among microbial classes can be found in the Supplementary Sections 1.2 and 1.3.

\section*{Deep Learning Analysis}

To verify the suitability of the AGAR dataset for building deep learning models for image-based microorganism recognition, and to set a benchmark for the task, we evaluated the performance of the two neural network architectures for object detection: Faster R-CNN~\cite{bib:Faster2015} and Cascade R-CNN~\cite{bib:Cascade2018}, with four different backbones: ResNet-50~\cite{bib:resnet2016}, ResNet-1010~\cite{bib:resnet2016}, ResNeXt-101~\cite{bib:resnext2017}, and HRNet~\cite{bib:hrnet2020}. The implementation from the MMDetection~\cite{bib:mmdetection2019} toolbox was used. The backbones’ weights were initialized using models pre-trained on ImageNet~\cite{bib:imagenet2009}, available in the \textit{torchvision} package of the PyTorch library. More technical details are provided in Supplementary Section 4.1.

The quantitative results for each subset of the AGAR dataset analyzed separately are reported, as well as the results for models trained on the whole dataset. Each subset was split randomly into approximately 75\% samples, which constituted the training set, and 25\% for the validation and testing set. Due to the relatively high resolution of all images, the samples were divided into patches of 512×512 px to avoid scaling down, which would make the recognition of small colonies more challenging. If necessary, zero paddings were applied to get a proper aspect ratio. As the goal of this study was to establish a baseline for the microbe detection task, the neural network architecture was not modified, thus the patches were resized to match the default input layer size. More details on the data split and patch preparation can be found in Supplementary Section 2.1.

Several approaches have been undertaken for data augmentation, including the addition of Gaussian blur, salt and pepper noise, changing the image color space to LAB and HSV, preparing histogram equalization, rotating images in the \mbox{[-45, 45]} degree range, or cropping around the annotated bounding boxes, but always keeping a visible microbe object. However, the best results were obtained through splitting the image into patches and normalizing it using means and standard deviation per channel values, the same as for the Common Object in Context (COCO) dataset~\cite{bib:COCO2014}.

The evaluation metric for colony detection was based on the mean Average Precision (mAP) previously established for the COCO competition~\cite{bib:COCO2014}. The Average Precision (AP) is calculated as the average value of precision on the precision-recall (PR) curve for a given Intersection over Union (IoU) threshold. IoU describes the level of overlapping between truth and predicted bounding box. The mean value of AP for different IoU thresholds and/or classes gives the mAP. To measure the effectiveness of the colony counting, two separate metrics were provided, namely Mean Absolute Error (MAE), its cumulated variation (cMAE), and symmetric Mean Absolute Percentage Error (sMAPE), and evaluated for different models on every data subset. Our post-processing of the predicted bounding boxes is described in detail in the Supplementary Section 2.2, while the calculations of all used metrics (mAP, MAE, cMAE, sMAPE) in the Supplementary Sections 3.1 and 3.2.

\begin{figure*}[!tb]
\centering
\includegraphics[width=.8\textwidth]{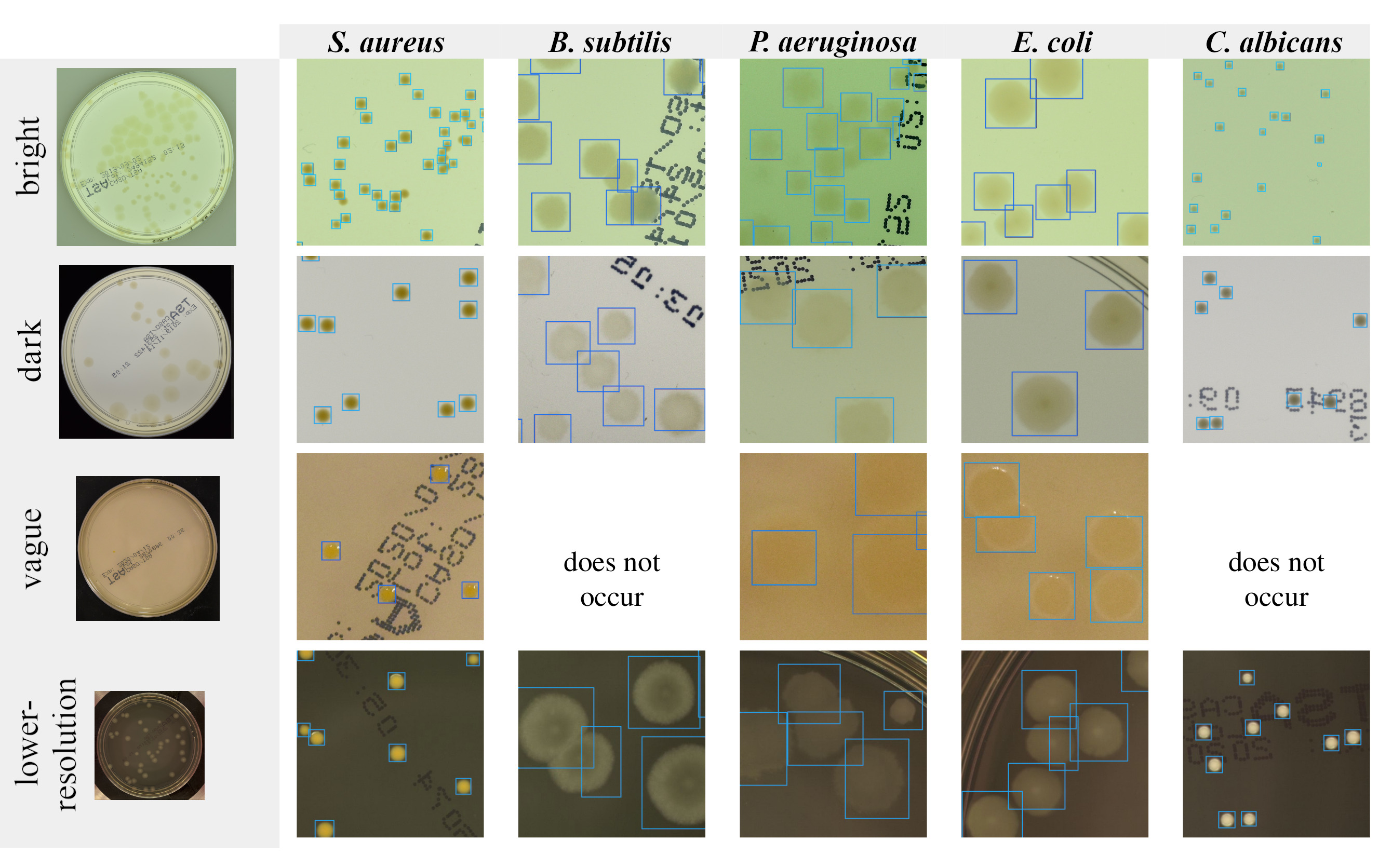}
\caption{\label{fig:boxed}\textbf{Examples of Cascade R-CNN with HRNet predictions for different types of microorganisms grown on TSA (Petri plates) captured with different image acquisition setups.} Entire plates preserving the photoscale are presented in the first column. The other columns represent 20×20 mm fragments of the plate. Well suited bounding boxes were found for organisms forming smaller, dense colonies, like \textit{S. aureus} and \textit{C. albicans}. In the case of \textit{P. aeruginosa} growing in large translucent colonies, some were omitted, especially in the \textit{vague} subgroup, where the edges of colonies could not be clearly defined. The \textit{vague} subgroup does not include \textit{B. subtilis} and \textit{C. albicans} at all.}
\end{figure*}

\begin{figure*}[!tb]
\centering
\includegraphics[width=.9\textwidth]{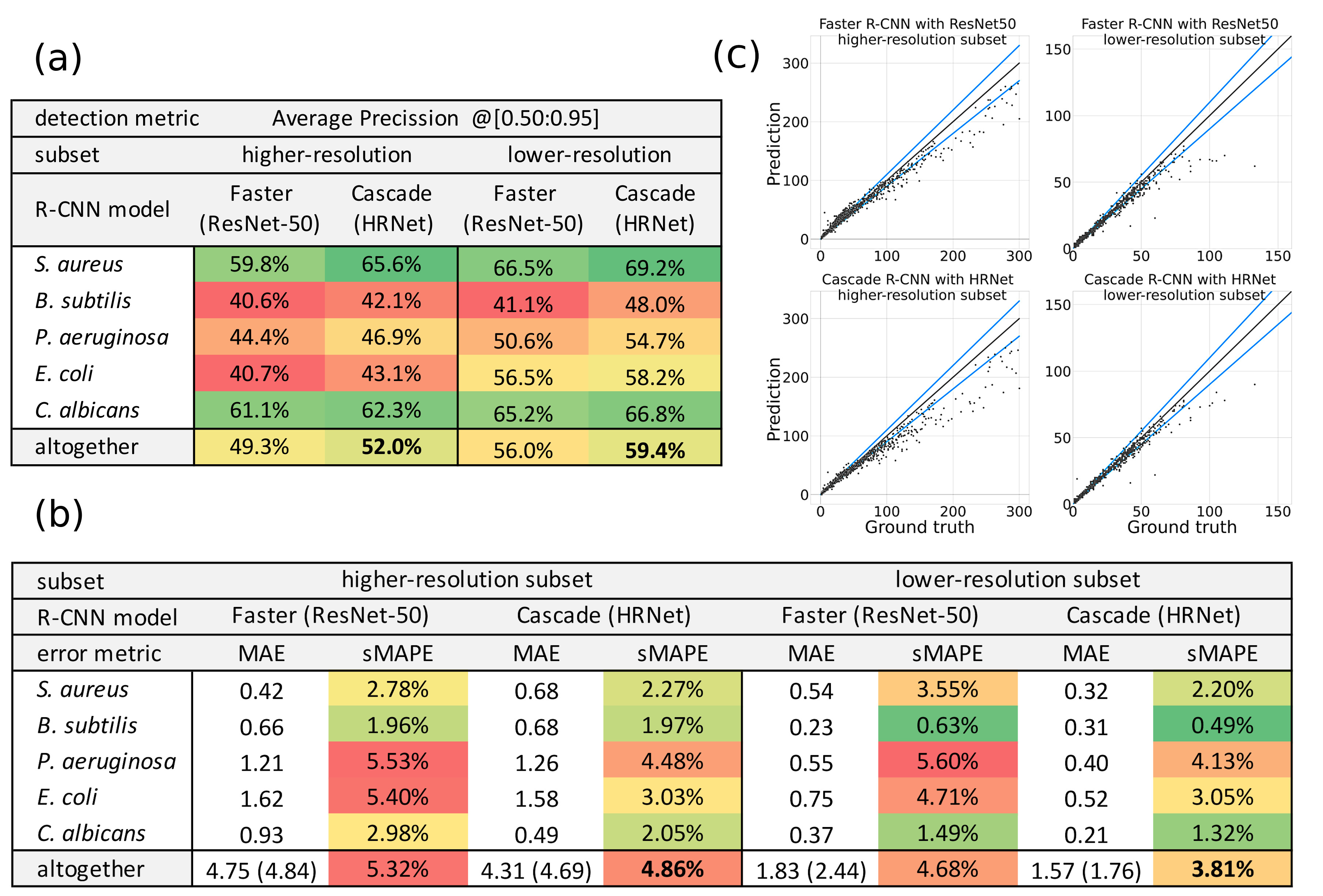}
\caption{\label{fig:results}\textbf{Overall results for colony detecting and counting tasks.} (a) Average precision for the two different models obtained on the \textit{higher}- and \textit{lower-resolution} subsets: the Cascade R-CNN model with HRNet backbone gives the highest mAP. (b) MAE and sMAPE metrics for Faster R-CNN and Cascade R-CNN calculated separately for each class of microbes and jointly (including cMAE in brackets). (c) Microbial colony counting performance -- the number of predicted colonies versus ground truth (each point represents the result for a single plate).}
\end{figure*}

The analysis for each of the major subsets of the AGAR dataset (\textit{higher}- and \textit{lower-resolution}) was performed independently. The vague subset was excluded from the studies. It was later used as a different domain representation in the studies on the models’ ability to generalize to new acquisition setups, see Supplementary Sections 4.7 and 4.8 for details. The results for the baseline model (Faster R-CNN with ResNet-50), along with more complex, multi-stage detectors (Cascade R-CNN with HRNet), are presented in Fig.~\ref{fig:results}. The visualisation of predictions made by Cascade R-CNN with HRNet backbone is shown in Fig.~\ref{fig:boxed}. Detailed descriptions of results achieved by other tested detectors are placed in Supplementary Sections 4.2 and 4.3.

Our main outcomes in terms of mAP are presented in Fig.~\ref{fig:results}~(a). In general, Cascade R-CNN performs slightly better with mAP scores equal to 52.0\% for the \textit{higher}- and 59.4\% for the \textit{lower-resolution} subset, compared to Faster R-CNN, which achieved 49.3\% and 56.0\%, respectively. A comprehensive study showed that the increase of backbone complexity gives just slightly better results (see Supplementary Section 4.2), which suggests that, in this class of architectures, the model’s capability is reached and extending backbone networks further might not lead to much better performance.

The average precision calculated per microbe is higher for microbes forming smaller colonies, namely \textit{S. aureus} and \textit{C. albicans}. The rest of the studied microorganisms tend to aggregate or overlap, especially in low dilution samples (more colonies on the plate surface). Moreover, larger colonies tend to have blurred edges with lower contrast with regard  to agar substrate. This all leads to less accurate microbe detection in terms of mAP.
The mAP score measures the quality of detection using IoU, which is related to true and predicted bounding boxes overlapping. For the colony counting, the recognition of microbes is crucial, but the exactness of their location is of secondary importance. Therefore, the sMAPE and MAE metrics were used for tuning the algorithms applied in the post-processing stage to merge predictions for individual patches into the whole test image. The values of two parameters, namely classification probability threshold and NMS threshold for the modified soft non-maximum suppression algorithmm~\cite{bib:NMS2017}, were established on the training dataset and then used to eliminate  the excess of bounding boxes (e.g. appearing twice through the edges of neighbouring overlapped patches) for the test data.

The results of microbe counting are presented in Fig.~\ref{fig:results}(b). Overall, Cascade R-CNN with HRNet predictions are more accurate with sMAPE equal 4.86\% for \textit{higher}- and 3.81\% for \textit{lower-resolution} subsets. It is worth noting that the most lightweight of the considered models (Faster R-CNN with ResNet-50) is also very effective with 5.32\% and 4.68\% sMAPE for the same subsets. As these metrics do not include the error coming from microorganism misclassification, cMAE being a sum of MAE calculated for every microbe separately is also provided (in brackets in \textit{altogether} row). For cMAE, only properly recognized microbes are taken as a correct count. The small difference between MAE and cMAE (e.g. 1.57 vs 1.76 in case of Cascade with HRNet for \textit{lower-resolution} subset) proves the great ability of the models to distinguish between different species.

The selected models’ predictions for microbial counting on test AGAR subsets are presented in Fig.~\ref{fig:results}(c). The straight black lines represent perfect results, while the blue ones indicate a 10\% error to highlight the acceptable error range. The detectors tend to underestimate the exact number of colonies for more populated samples. This is largely due to the nature of the sMAPE metric, which weights errors inversely to the number of instances. Moreover, the AGAR dataset is dominated by plates with less than 50 colonies (see Fig.~\ref{fig:statistics}(c)), which impacts the values of the probability and NMS thresholds for post-processing. The extension for higher populated samples using double thresholding approach is described in Supplementary Section 4.6.

Although \textit{empty} and \textit{uncountable} plates come with no colony annotations, qualitative studies were carried out to evaluate the performance of the trained models.  Experiments on uncountable samples showed that many plates labeled by microbiologists as having more than 300 colonies were, in fact, within the countable range, as recognized by the detectors. As discussed above, the models are more accurate for samples with less than 50 colonies. However, they give very good estimates for plates with hundreds or even thousands of instances, correctly identifying single colonies in highly populated samples, with the maximum number of predicted colonies on one plate equal to 2782. It is worth noting that it takes seconds for the deep learning system, while it could take up to an hour in the case of manual counting.

The analysis of samples marked by microbiologists as \textit{empty} presented the great capability of deep learning methods for microbiological quality control. The detectors were able to recognize colonies difficult to see and missed by human specialists. Overlooked microbes were usually small and located either at the edge of a Petri dish or on a plate factory marking. This is a strong indication that human-AI interaction can lead to a higher quality of microbiological analyses.

Further discussion on empty and uncountable samples can be found in Supplementary Section 4.5. The complementary analysis of the impact of the models’ initialization parameters are provided in Supplementary Section 4.4. The detectors’ ability to generalize to different AGAR subsets with transfer learning technique is described in Supplementary Section 4.7. The models’ performance on every subset when trained on the whole dataset is provided in Supplementary Section 4.8.

\section*{Discussion and Conclusions}

There is a growing interest in applying computer vision algorithms to microbiological analyses, especially in industry and pharmaceutical branches, but many existing approaches are based on traditional image processing techniques~\cite{bib:Zhang2008, bib:Brugger2012, bib:Zhu2018, bib:A1, bib:C1}. Although deep learning methods have recently become more commonly used~\cite{bib:Nie2015, bib:Zielinski2017, bib:Talo2019, bib:Ferrari2015, bib:Ferrari2017, bib:Savardi2018}, the proposed procedures are usually not end-to-end solutions and cannot be used independently to process an image of an entire Petri plate. The lack of effective DL-based approaches for microbiological purposes possibly results from the poor availability of huge datasets needed for deep neural network training. To fill this gap, the AGAR dataset is introduced with 18~000 photos of agar plate cultures (Petri dish) and 336~442 annotated colonies of five microorganisms (staphylococci, Gram-positive and negative bacilli, and yeast) most commonly used according to the Pharmacopoeia guidelines.

An earlier study by Ferrari et al.~\cite{bib:Ferrari2015, bib:Ferrari2017, bib:Savardi2018} presented the creation of the MicrobIA, a smaller dataset composed of about 29~000 labeled segments (fragments of the Perti dish with single agglomerates) of bacterial colonies grown on blood and chromogenic agar plates. The segments were extracted with the WASPlab system from photos of whole Petri plates and manually assigned to one of seven classes (segments containing from 1 to 6 colonies and outliers). The MicrobIA dataset was then used to train a convolutional neural network to classify images into one of the six colony counts or the outliers category (bubbles, dust or dirt on the agar). Authors reported the per-colony error of 28\% measured on the individual segments~\cite{bib:Ferrari2017}. Another study using the MicrobIA dataset~\cite{bib:Andreini2020} explored the possibility of image generation for agar plates’ segmentation and investigated the domain shift problem. It was proved that synthetic data can be applied to train a deep neural network for agar plate image segmentation.

In contrast, our study demonstrates the possibility of exploiting deep learning-based detectors to build end-to-end solutions for multiclass microbial colony recognition and counting. The provided benchmarks of the chosen DL models prove that the AGAR dataset can be used to build robust models, which adjust well to real data achieved in various acquisition setups, resulting in different illumination and resolution of a Petri plate photo. The exhaustive analysis of eight different deep neural networks for object detection was performed on two AGAR subsets referred to as \textit{higher}- and \textit{lower-resolution}. The best performing model used in our study (Cascade R-CNN with HRNet backbone) achieved the low counting error of 4.92\% and 3.81\%, respectively on both subsets. This is a significant improvement over previous reports describing the microbial colony counting with computer vision algorithms~\cite{bib:Ferrari2017}. In the case of detection, the mAP scores within the range from 49.3\% to 59.4\% were achieved for different detectors, which is an excellent result compared to other reports (44.6\% for Cascade R-CNN and 36.7\% for Faster R-CNN [16]) done with the same architectures on the COCO dataset.

In summary, the selected R-CNN models perform very well in detecting microbial colonies. This is likely because the detected instances have similar shapes and all species of microbes are well represented in the training data. Moreover, the results obtained with base Faster R-CNN and more complex Cascade R-CNN do not differ much. For that reason, exploring different state of the art AI approaches, such as Transformers~\cite{bib:Zhu2020}, seems to be the right future direction for the implementation of deep learning in microbiology.

There is a visibly increasing demand for artificial intelligence in numerous human activities, including social interactions, industry, agriculture, and medicine. The AGAR dataset, which compiles a huge variety of Petri dish culture photos, is a great data source for further exploration. The use of generative models on AGAR might, for example, improve the performance of colony detection on a wider variety of images from other domains, as well as with a larger number of microbial species. Additionally, the large variety in the distribution of colony sizes shows the potential of the AGAR dataset for detailed studies of colony growth dynamics, which can, for example, help us better understand antibiotic resistance of bacteria.

\bibliography{main}

\section*{Acknowledgements}
Project “Development of a new method for detection and identifying bacterial colonies using artificial neural networks and machine learning algorithms” is co-financed from European Union funds under the European Regional Development Funds as part of the Smart Growth Operational Program. Project implemented as part of the National Centre for Research and Development: Fast Track (grant no. POIR.01.01.01-00-0040/18).

\section*{Author contributions statement}

SM, JP, TB, AH and AP contributed to the DL analysys code. GG and JR prepared microbiological samples, captured photos of standard agar plates culture and annotated them. AL and TG prepared an application to annotate images. SM analyzed and interpreted the AGAR datasets statistics. SM and JP planned the experiments and developed DL-based methodology to identify and count microbial colonies. JP and SM performed calculations and analysed the results of the experiments. ZDK led the microbiologist team, while TG led the DL team during the whole project. TG and ZDK supervised the project.

All authors provided critical feedback and helped shape the research, the analysis and the manuscript.

\end{document}


\maketitle

\section{The AGAR dataset in details}
\label{sec:Dataset}

The AGAR is a color image dataset created for developing methods of microorganism colonies detection and counting. It contains 18~000 annotated photos of Petri dishes, which gave 336~442 annotated colonies in total. The images were taken under diverse lighting conditions with two cameras. More information about microbial colonies inoculation, and data acquisition setups can be found in Section~\ref{subsec:colonies_growth}, and Section~\ref{subsec:Data_specifications}. The image annotations procedure is described in detail in Section~\ref{subsec:Image_annotations}.

\subsection{Microbial colonies growth conditions}
\label{subsec:colonies_growth}

Based on European, American, and Japanese Pharmacopoeia~\cite{bib:M1, bib:M2, bib:M3} regulations and ATCC guidelines~\cite{bib:M15}, \textit{Staphylococcus aureus} subsp. \textit{aureus} Rosenbach FDA 209 - ATCC 6538, \textit{Bacillus subtilis} subsp. \textit{spizizenii} NRS 231 - ATCC 6633, \textit{Pseudomonas aeruginosa} R. Hugh 813 - ATCC 9027, \textit{Escherichia coli} Crooks - ATCC 8739, and \textit{Candida albicans} 3147 - ATCC 10231 strains were used for training and testing deep learning algorithms. Microorganisms have been purchased from American Type Culture Collection (LGC Standards, Poland) and were stored at -70\textsuperscript{o}C in Trypticase Soy Broth (TSB, Becton Dickinson and Company, Cockeysville,  MD,  USA)  supplemented with  20\%  glycerol  (U.S.  Merck  Corporate  Headquarters, Kenilworth, NJ, USA). For the experiment, strains were refreshed on Trypticase Soy Agar (TSA, Becton Dickinson and Company, Cockeysville, MD, USA) at 37\textsuperscript{o}C for 18 h. A series of 10-fold dilutions were made in Ringer's fluid of the initial culture at a density equivalent to 0.5 McF (Densimat Densitometer, bioMerieux SA, France). From selected dilutions, 100$\mu$l were cultured onto TSA medium in five technical replicates. Cultures were inoculated using sterile 6 mm diameter glass beads on a laboratory rocker (Multi Bio-3D Sunflower, Biogenet, Poland) for 40 s. The beads were then removed and the microbial cultures were incubated at 37\textsuperscript{o}C for 18-24 hours.

\subsection{Data acquisition}
\label{subsec:Data_specifications}

\begin{figure}[!hbt]
\centering
\includegraphics[width=\textwidth]{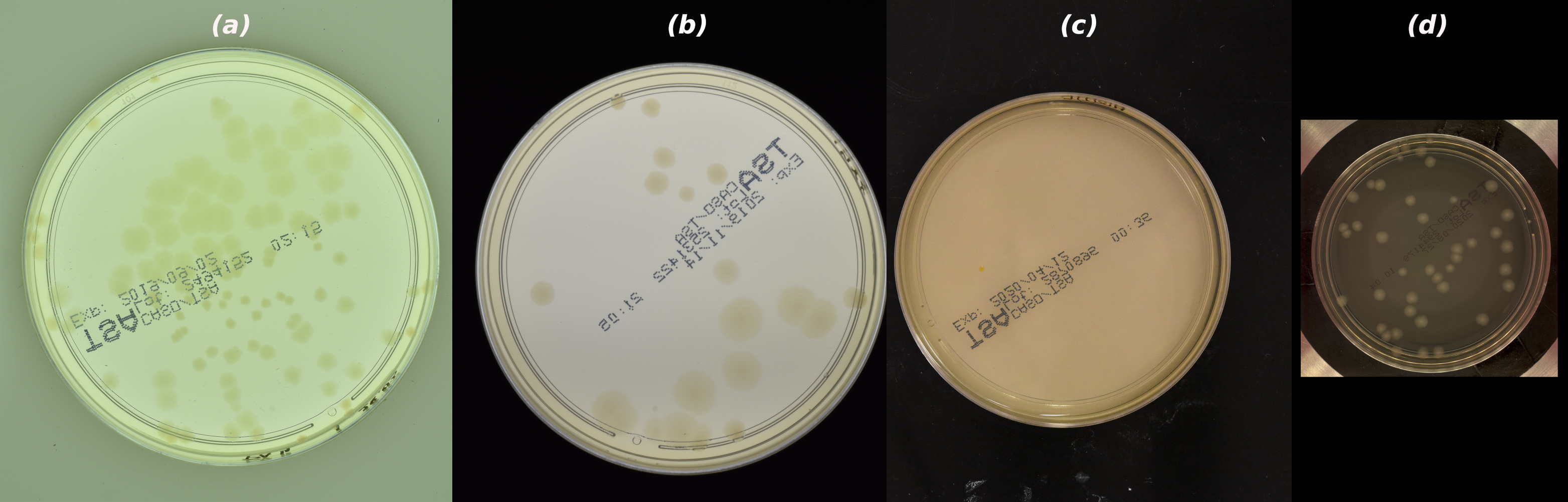}
\caption{\label{fig:plate_background} Example photos of agar plate cultures of AGAR dataset for different image acquisition setups: (a) \textit{bright}, (b) \textit{dark}, (c) \textit{vague}, and (d) \textit{lower-resolution}.}
\end{figure}
A simple image acquisition system consists of a Nikon D3500 camera with 60~mm lens and 24~Mpx CCD and a plexiglass stand for Petri dishes, with LED source light mounted below. Data collected with this setup is labeled as \textit{higher-resolution} (4~000 x 6~000~px). In order to provide some diversity, small changes in the setup during the data acquisition process were introduced. Thus, three subgroups of \textit{higher-resolution} photos are distinguished: \textit{bright}, \textit{dark}, and \textit{vague} (see Fig.~\ref{fig:plate_background}). The former two were made with the whole setup closed in a box to eliminate the influence of ambient light. The difference between them lies in a color of plexiglass used for a stand---white for \textit{bright} subgroup, or black for \textit{dark} one. The last subgroup, \textit{vague}, was exposed to available light, which resulted in low contrast images, hard to annotate even by a professional microbiologist.

A different setup was used to collect the part of the dataset referred to as \textit{lower-resolution} (2048 x 2048~px). The system includes a holder designed to mount a Petri dish on (visible on photos), a monochrome camera (IDS UI-5370CP-M-GL with 4.19~Mpx resolution), and a set of adjustable light sources below and above a holder (their parameters were changed during data acquisition process). RGB images are obtained as a composition of three monochromatic photos made with different optical filters, corresponding to red, green, and blue colors. Example photos of agar plate cultures taken under different conditions are presented in Fig.~\ref{fig:plate_background}.

\subsection{Image annotations}
\label{subsec:Image_annotations}

The design of a robust annotation pipeline was critical due to the desire to label 18~000 Petri dish photos, which led to over 330~000 labelled colonies. The image acquisition setup was equipped with a digital counter controlled by Raspberry Pi to display a sample identification (ID) number. This allowed us to verify if the appropriate photos were uploaded for the corresponding entry in the database. Please note that images are automatically cropped (based on their histograms) during the first stage of preprocessing to keep only a part with a Petri dish (and stored in this form in AGAR dataset), however, in some cases a part of the counter can be still visible.

\begin{figure}[!htb]
\centering
+\includegraphics[width=.4\textwidth]{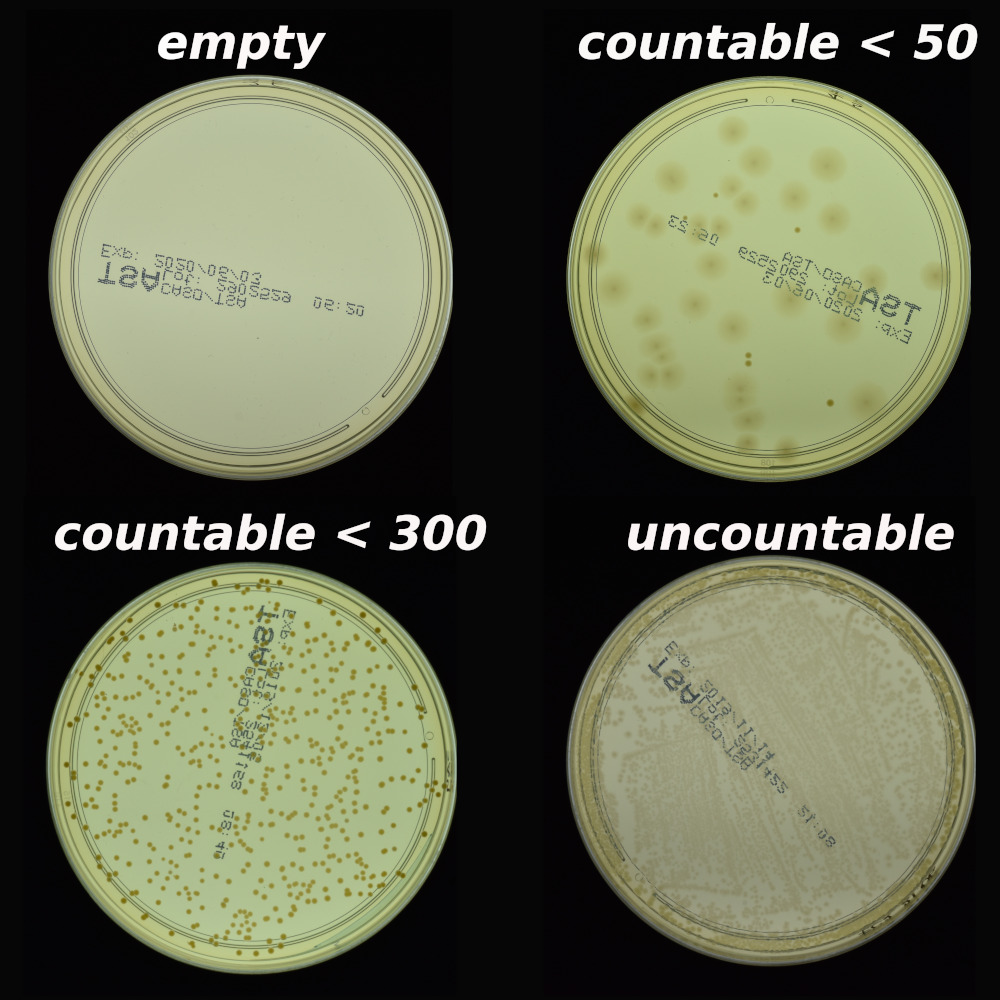}
\caption{\label{fig:count_probes} Examples of images of agar plate cultures with different number of colonies.}
\end{figure}

A web application was developed for microbiologists to upload and annotate agar plate culture photos. Each sample recorded in the database is marked as \textit{countable}, \textit{uncountable}, or \textit{empty}. There are some indications~\cite{bib:Tomasiewicz1980, bib:Sutton2011} of the optimal number of colonies to perform manual counting. Based on them, samples with more than 300 colonies are treated as \textit{uncountable}. Please note, however, that some samples may also be marked as \textit{uncountable} if there were problems with clear identification of colonies (e.g. due to low contrast in the \textit{lower-resolution} subset). Similarly, for samples with colonies number range between 50 and 300 instances, some colonies agglomerated, and it was difficult to identify their boundaries (see Fig.~\ref{fig:count_probes}). In case of samples identified as \textit{countable}, microbiologists annotated every colony with its location (by drawing a bounding box) and class. There are 7 possible classes: 5 microorganisms (\textit{S. aureus, B. subtilis, P. aeruginosa, E. coli, C. albicans}), defects (crack marks on an agar surface), and contamination (microbial contamination like environmental microorganisms and sometimes fungi).

The annotations are stored in JSON files provided for every sample, with the same basename as the corresponding image file. They all share the same data structure presented in Listing~\ref{lst:annotations}. Each file contains a series of fields, including: the background category name (\textit{bright}, \textit{dark}, \textit{vague}, or \textit{lower-resolution}), the list of names of microbe species that were inoculated on a Petri dish (note that sometimes they have not grown), the total number of colonies (excluding defects and contamination), the list of bounding boxes, and the sample ID. Provided bounding boxes give information about pixel coordinates of the top-left corner, its width and height, microbe class, and annotation ID. Please note that in the case of \textit{uncountable} samples $colonies\_number$ key gets -1 value. The file structure is presented in the Listing~\ref{lst:annotations}.
\begin{lstlisting}[caption={The data structure of a single annotation JSON file.}, label={lst:annotations}, captionpos=b, basicstyle=\small]
{   "background": str,
    "classes": [str],
    "colonies_number": int,
    "labels": [
                {
                 "id": int,
                 "class": str,
                 "height": int,
                 "width": int,
                 "x": int,
                 "y": int
                }, ...
              ]
    "sample_id": int }
\end{lstlisting}
For convenience, annotations are also provided in the well known COCO format~\cite{bib:COCO2014}, with an additional key describing background category.

\subsection{Statistical data analysis}
\label{subsec:Statistical_analysis}

\begin{figure}[!tb]
\centering
\includegraphics[width=.51\textwidth]{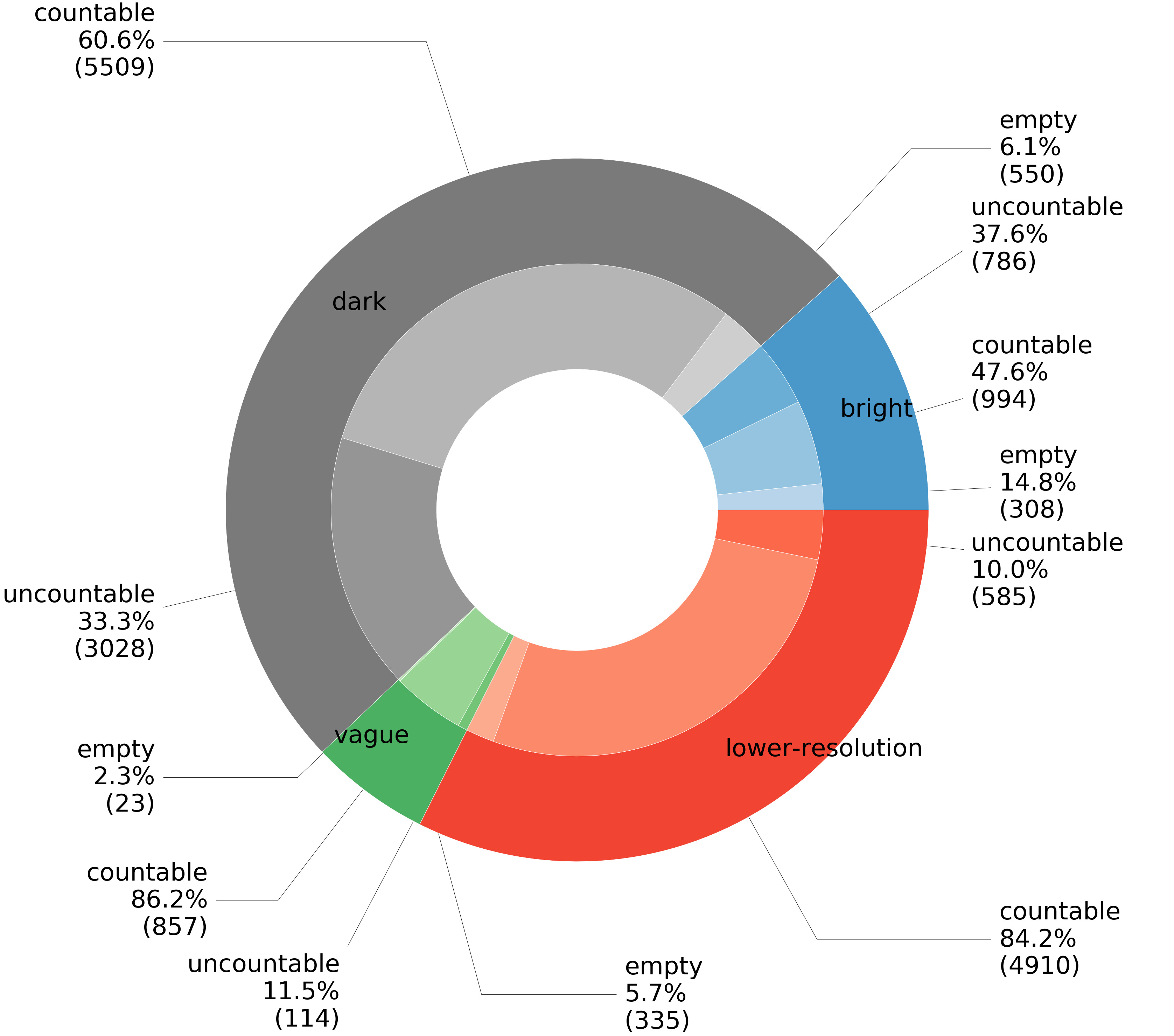}
\includegraphics[width=.47\textwidth]{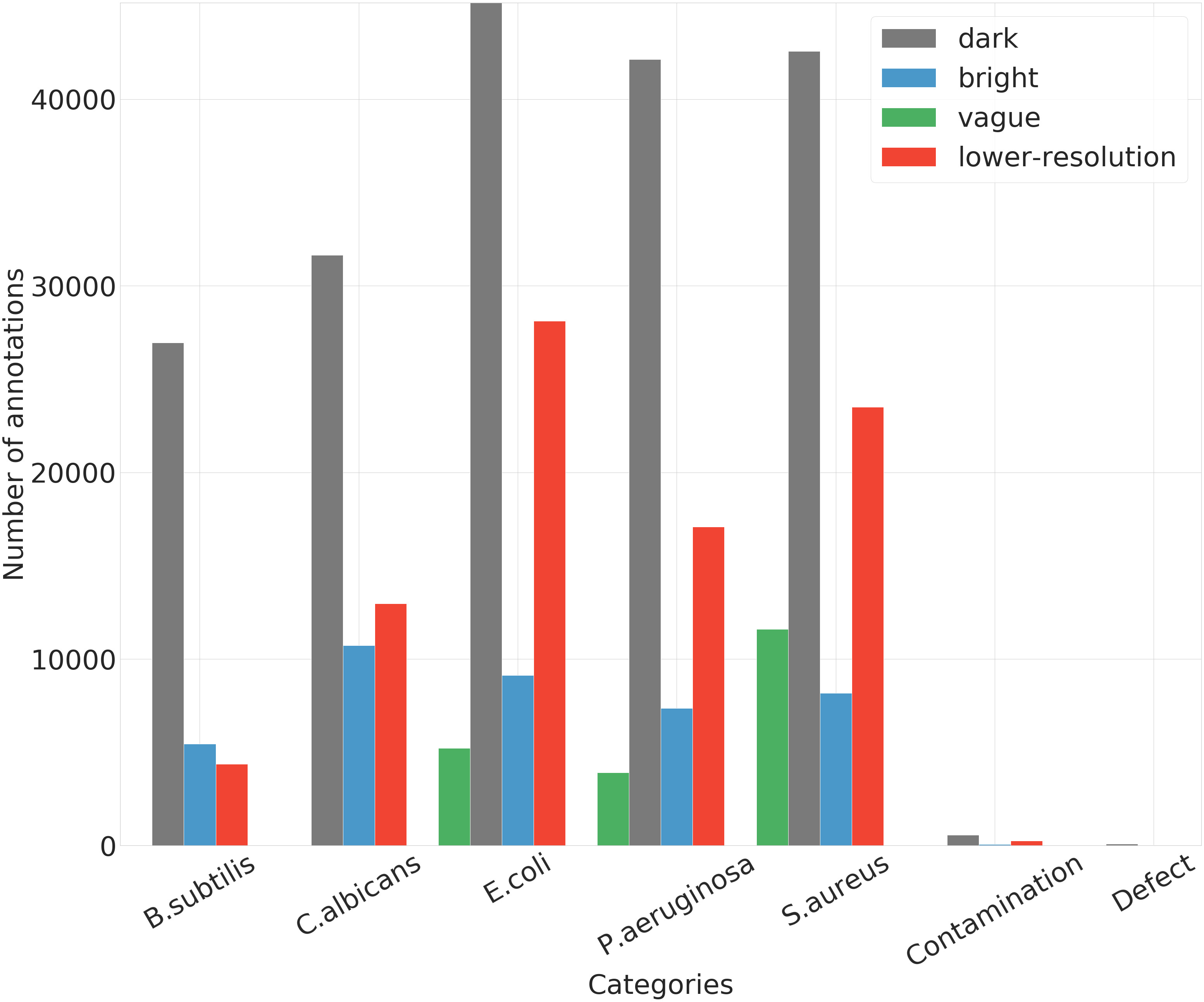}
\caption{\label{fig:pie_hist} (left) The distribution of samples over different subgroups. (right) The number of instances per category for all 5 microbe types and additional two classes -- defects and contaminations.}
\end{figure}

The summary of AGAR dataset statistics showing the distribution of images over different background categories is presented on a pie chart in Fig.~\ref{fig:pie_hist}(left). Each color represents different background category, while different shades indicate sample classes (\textit{empty, countable}, or \textit{uncountable}). The samples from \textit{dark} subgroup constitute about half of the whole dataset and they include photos with the best quality in terms of contrast between grown microorganisms and an agar surface. At the very beginning of the data collection process, the full range of dilutions was used to collect the variety of samples, as in the case of \textit{higher-resolution} subgroup. However, then a low dilution level was used less often to reduce the number of \textit{uncountable} plates. That is why the \textit{lower-resolution} group, acquired later, includes much less of them than the other subgroups.

In general the AGAR dataset achieves a good balance in the number of instances of different microbe species, which is important for building a robust deep learning model. However, given subgroups may be partially unbalanced as presented in Fig.~\ref{fig:pie_hist}(right). The \textit{vague} subgroup does not include \textit{B. subtilis} and \textit{C. albicans} at all, but it has more than 80\% of photos with two different microbe species, in contrast to 15\% for \textit{higher-resolution} (excluding \textit{vague}) and 40\% for \textit{lower-resolution}. It is also not as numerous as other subgroups, because images from \textit{vague} subset are difficult to annotate. {\it B. subtilis} is also underrepresented in the \textit{lower-resolution} subset, though, it is still enough to train a detector to recognize this species.

\begin{figure}[!hbt]
\centering
\includegraphics[width=.55\textwidth]{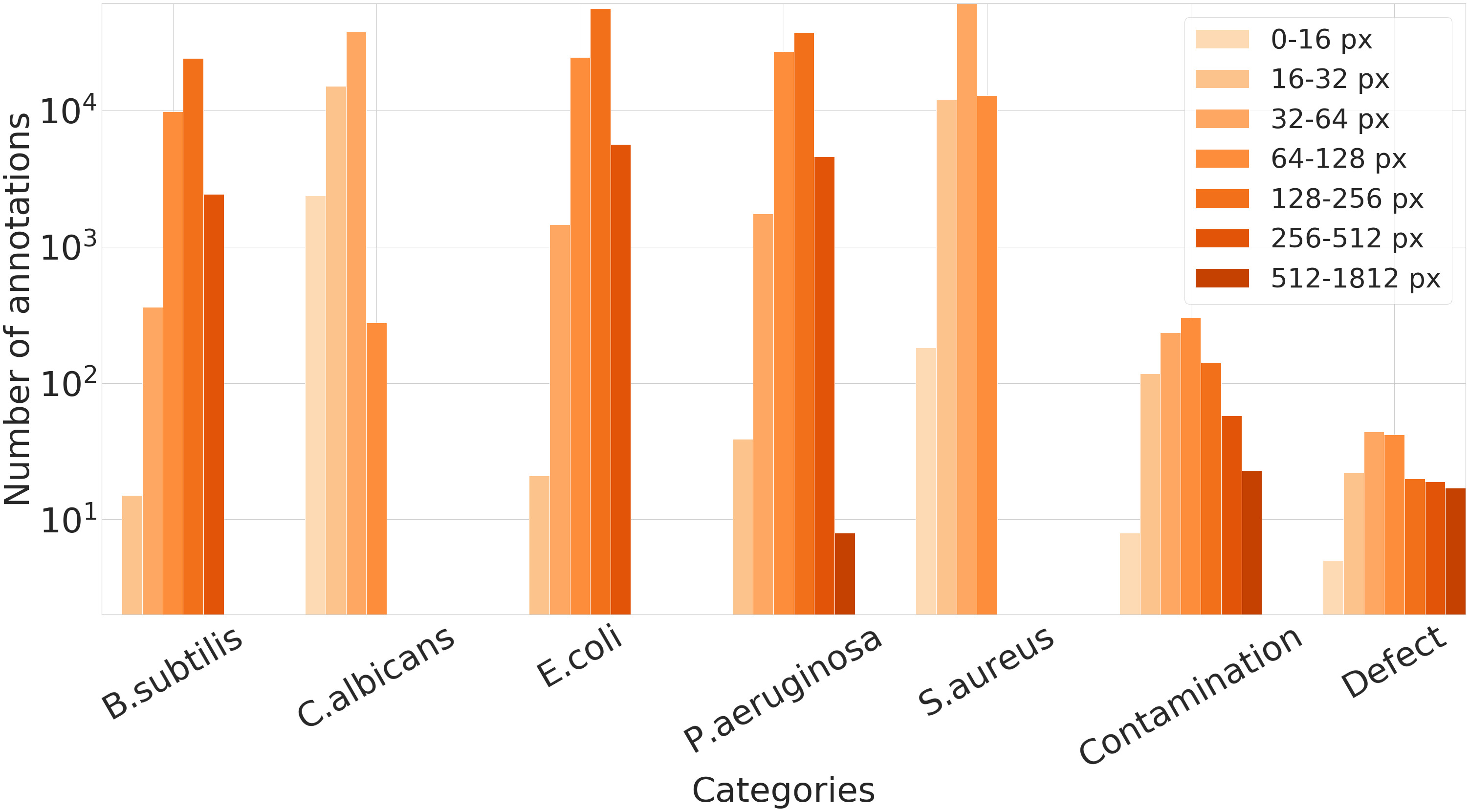}
\caption{\label{fig:area}The number of instances per category for all annotated instances grouped by square root of bounding box area.}
\end{figure}

Fig.~\ref{fig:area} shows the size variability of annotations per category for the whole dataset. Two basic box size distributions can be distinguished: 0-128 px for \textit{C. albicans} and \textit{S. aureus}, and 16-512 px for \textit{P. aeruginosa}, \textit{B. subtilis}, and \textit{E. coli}. In total (excluding defects and contamination), there are 154~630 bounding boxes with squared root of area bellow 128~px, 180~173 within 128-512~px, and 1~639 above 512~px. This wide range of sizes makes the detection task more challenging because models have to be flexible enough to handle the variety of instances' dimensions.

\begin{figure}[!hbt]
\centering
\includegraphics[width=.4\textwidth]{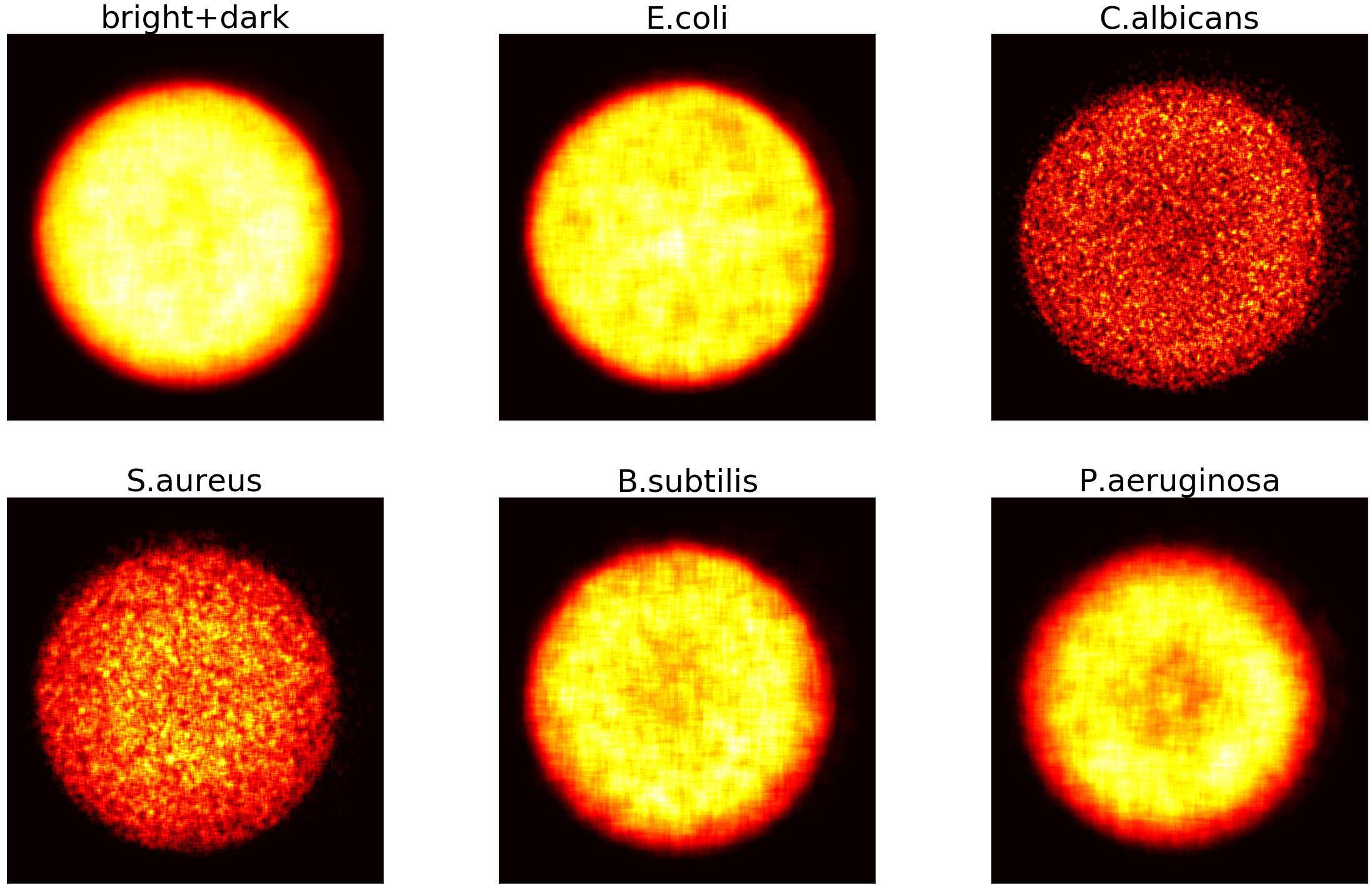}
\caption{\label{fig:sun_bacteria} Heatmaps of all annotations at one picture for \textit{higher-resolution} subgroup. Brighter colors indicates greater crowd in a given place in the picture. All results was normalized using maximum of a given subset (for each microbial species).}
\end{figure}

Another interesting aspect is the distribution of growing colonies on the plate, as it is presented in Fig.~\ref{fig:sun_bacteria} (only for \textit{higher-resolution} background category, excluding \textit{vague} samples). \textit{C. albicans} and \textit{S. aureus} form smaller colonies, which are more spatially separated on a Petri dish. The rest of considered species mainly cover a middle ring, which can be related to the method of microbes inoculation on an agar medium. 

\begin{figure}[!htb]
\centering
\includegraphics[width=.75\textwidth]{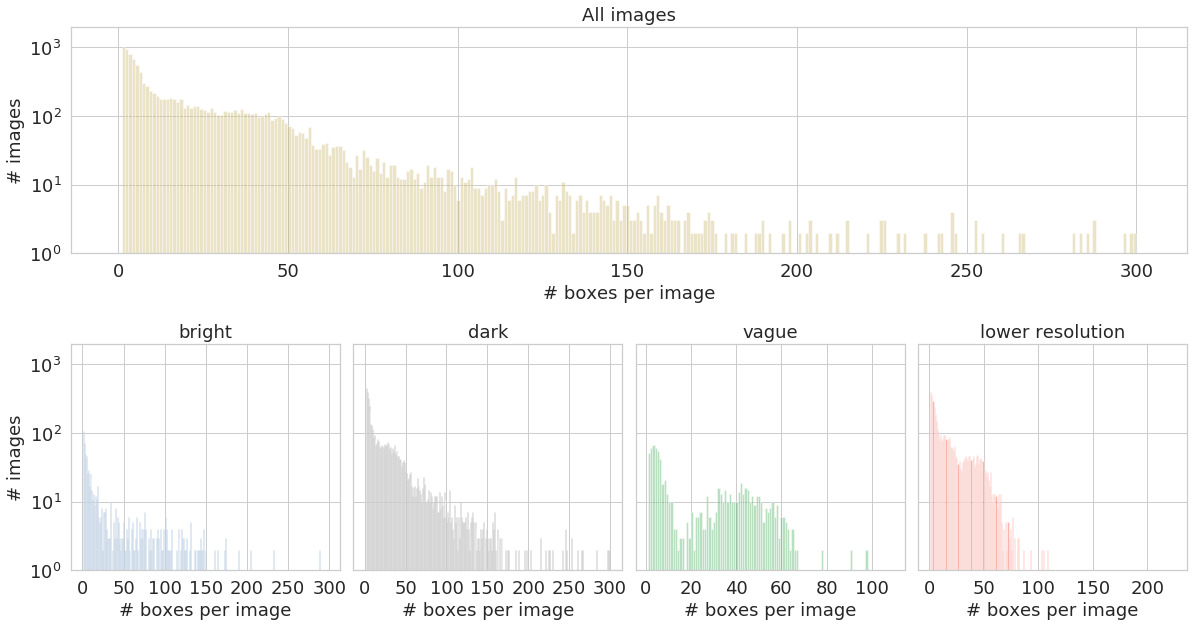}
\includegraphics[width=.75\textwidth]{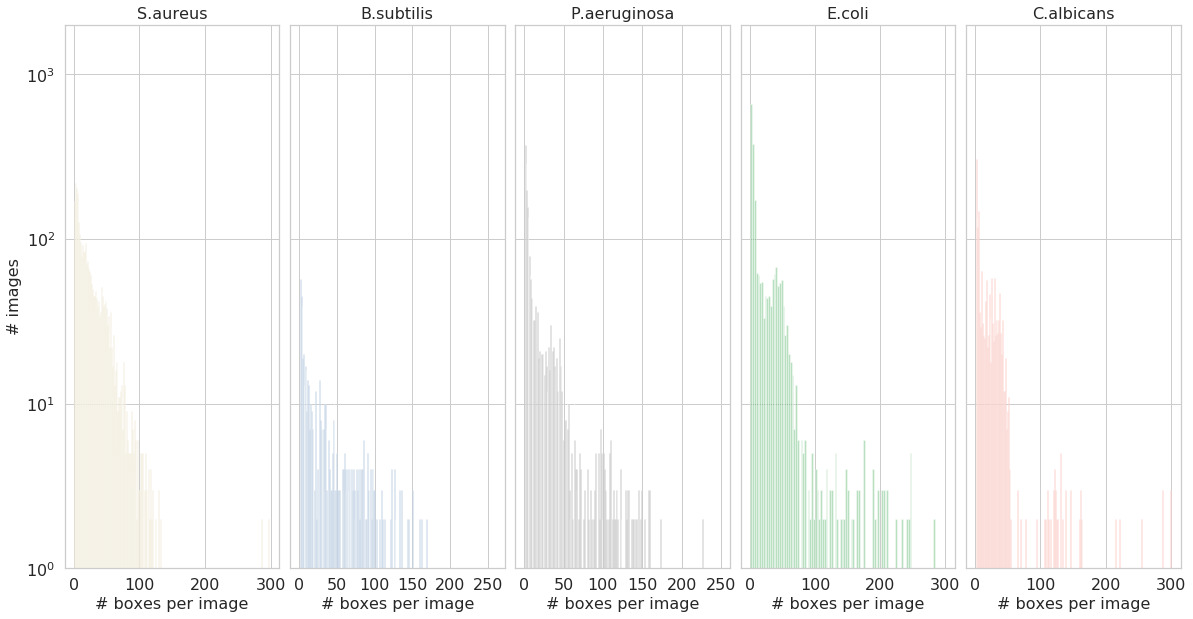}
\caption{\label{fig:count_histograms}Histograms of the number of annotated instances
per image, calculated for: the whole dataset, different subsets, and different microbial species.}
\end{figure}

As mentioned before, the \textit{countable} class includes samples with less than 300 colonies. The histograms shown in Fig.~\ref{fig:count_histograms} present the distribution of \textit{countable} samples in terms of the number of annotated colonies. The histograms of number of annotated colonies per image divided into microbial species are also included. In general, the interquartile range for the whole dataset is between 4 and 38 colonies per image. In the case of \textit{B. subtilis}, higher number of instances can be observed outside this range. Also, the distribution of \textit{vague} part of dataset shows slightly different behaviour where median is a bit higher.

\newpage
\section{Data processing}
\label{sec:processing}

AGAR images are very large in size compared to regular datasets like COCO~\cite{bib:COCO2014}. The original image resolution in the dataset ranges from about 2048 × 2048 to about 4000 × 6000 pixels, while most images in common datasets (e.g. COCO) are no more than 1000 × 1000 pixels. The annotations prepared by microbiologists were made on resized images, so, during pre-processing, the pixel positions of bounding boxes have to be recalculated. Also, during pre-possessing, the visible parts of electronic counter (which helped us to distinguish the samples from each other during the photo collection process) were cut out from the images. Nevertheless, such big images cannot in raw state go into neural network, thus pre-processing to split images into smaller parts should be applied first. Our pre-processing methods are described in Subsection~\ref{subsec:preprocessiong}. Post-processing methods that allowed us to count colonies from entire photos are presented in the Subsection~\ref{subsec:postprocessing}.

\subsection{Data pre-processing}
\label{subsec:preprocessiong}

In order to ensure that the training and test data distributions approximately match, 3/4 of the original images were randomly selected as the training sets and the rest were kept as validation sets for each background category separately. Please note that the same image subset is used for validation and test sets, however patches for the both sets are generated in different manner, as explained below. The list of image IDs of all original images used for training and validation set is publicly provided, but exact prepared patches are not.

\begin{figure}[!ht]
\centering
\includegraphics[width=.5\textwidth]{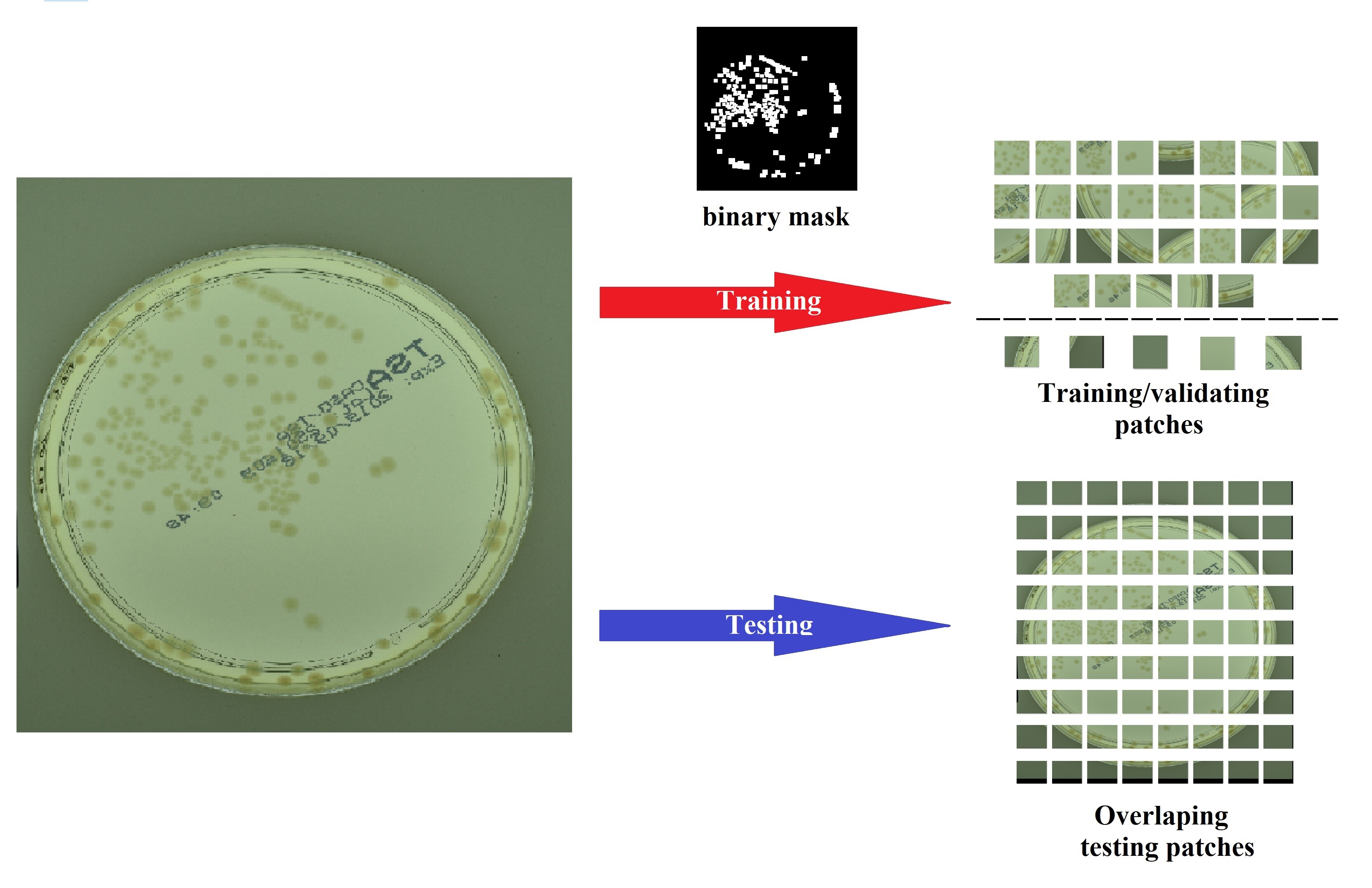}
\caption{\label{fig:patches}Example random patches for training/validation pipeline and overlapping testing patches. A dashed line separates empty patches (containing no colonies) from those that are occupied for training/validating mode.}
\end{figure}

To control the balance between empty patches and the ones containing microbes, first only training patches with at least one colony are generated and then the appropriate number of empty ones (few percent of number of generated patches with colonies) are added. The algorithm to divide images consists of a few subsequent steps. Firstly, a binary mask of all area of bounding boxes present on given image (white pixels make up the colony instances, black---the background) is generated. In the following steps, one box is chosen randomly, and a patch of size 512x512 containing the whole box (the left-top corner selection is also random) is cut out, and all other boxes (if any) on the patch are also marked as chosen. The above steps are repeated until every box at the image is marked. This procedure is illustrated as the red path in Fig~\ref{fig:patches}.

On the other hand, extracting test data patches is done without exploiting annotations. A sliding window with an overlap being equal to 1/8 of the width of the patch is used to get adjacent images parts (blue path in Fig~\ref{fig:patches}). The overlapping is necessary to get the final prediction for the whole image, and not to miss any colonies which accidentally lie in a division area. Removing redundant predictions is done in the post-processing of output given by a neural network.

\subsection{Data post-processing}
\label{subsec:postprocessing}
The post-processing consists of gathering together predicted bounding boxes on every patch belonging to one image and recalculating their exact position (rescale and use an appropriate offset taking into account given patch position relative to the whole image). After that, excessive boxes are filtered out using: (1) the probability of the detected microbe class and (2) the variation of the Soft Non-Maximum Suppression (NMS) algorithm~\cite{bib:NMS2017}. Our modified version of soft NMS based not on the biggest bounding box confidence level, but on the biggest area of prediction. The thresholds for both filters are tuned on training data (using simple grid search for each model separately) in a way to minimize metrics for microbe counting, defined in Section~\ref{sec:metrics}.

\section{Models evaluation metrics}
\label{sec:metrics}

Two stages of the task are considered: detection -- localization and classification (Section~\ref{subsubsec:detection_metrics}), and then counting the colonies of microorganisms (Section~\ref{subsubsec:counting_metrics}). As an evaluation metric for colonies detection, we rely on the mean Average Precision (mAP) established for the COCO competition~\cite{bib:COCO2014}. The Average Precision (AP) is the precision integrated over the whole recall range for precision-recall (PR) curve at given Intersection over Union (IoU) threshold. IoU describes the level of overlapping between truth and predicted bounding box. The mean value of AP for different IoU thresholds and/or classes gives mAP. To measure the effectiveness of colony counting we provide three separate metrics, namely Mean Absolute Error (MAE), its cumulative extension called cMAE, and Symmetric Mean Absolute Percentage Error (sMAPE) evaluated for different models on every data subset.

\subsection{Detection metrics}
\label{subsubsec:detection_metrics}
Two neural network architectures for two-stage object detection are evaluated to recognize and localize microbial colonies. Their performance is mainly measured using AP@[0.50:0.95] (mAP over different IoU thresholds, from 0.5 to 0.95, step 0.05).

IoU measures how well predicted bounding boxes fit the real location of an object. It is defined as
\begin{equation}
    \mathrm{IoU} = \frac{\mathrm{area\,of\,intersection}}{\mathrm{area\,of\,union}},
\label{eq:iou}
\end{equation}
where the intersection and union refer to true and predicted bounding boxes. To decide if a bounding box proposed by a model is valid or not, one should choose appropriate threshold value. For the value of selected threshold we can distinguish:
\begin{itemize}
    \item True Positive (TP) detection if IoU is greater than or equals the threshold value,
    \item False Positive (FP) detection if IoU is less than the threshold value,
    \item False Negative (FN) detection if a model fails to detect the proper object,
    \item and True Negative (TN) in case when model correctly did not predict any object.
\end{itemize}

With the above definitions, precision and recall are given by expressions
\begin{equation}
\begin{split}
    \mathrm{Precision} &= \frac{\mathrm{TP}}{\mathrm{TP} + \mathrm{FP}},\\
       \mathrm{Recall} &= \frac{\mathrm{TP}}{\mathrm{TP} + \mathrm{FN}}.
\end{split}
\label{eq:precision_recall}
\end{equation}

Precision shows how accurate predictions are, while recall measures how good a model is at finding objects. The precision-recall curves used in this study show the trade-off between precision and recall. Usually the increase of one metric causes the decrease of the other. High recall but low precision reflects the increase of the number of predicted bounding boxes, but classified incorrectly in most cases. On the other hand, high precision but low recall is just the opposite situation, a model returns very few results, but most of its predicted labels are correct.

Finally, the Average Precision for a given microbe class is calculated by integrating the are under the precision-recall curve (using e.g. the trapezoidal rule). It is usually done for different IoU thresholds. The mAP score is the mean over all thresholds, typically AP@[0.50:0.95] is considered~\cite{bib:COCO2014}, and/or all microbe classes.

\subsection{Counting metrics}
\label{subsubsec:counting_metrics}

Counting by detection is supposed to be the most natural approach to get the number of objects on an image. Even if localizing colonies itself is not necessary for this task, it is very useful to investigate the achieved results (especially to understand the cases in which model makes mistakes). To evaluate counting task the MAE, cMAE, and sMAPE were used.

MAE measures an average error between paired observations: ground truth count of all annotations and the number of predicted bounding boxes, for the whole image of a Petri dish. As the metric do not include the error coming from the misclassification of microorganisms, cMAE being a sum of MAE calculated for every microbe separately is also provided (given in brackets in altogether row in Tables~\ref{tab:mape_bright}-\ref{tab:mape_lower-resolution}). For cMAE only properly recognized microbes are taken as a correct count. sMAPE is similar to general MAE, but it weights particular samples errors inversely with the number of colonies on a plate. The intuition behind this metric is as follows. In a real world scenario, humans tend to perceive counts in the logarithmic scale. That means, that a mistake of 1 for an \textit{empty} dish might seem intolerable but the same mistake for a ground truth count of 50 might seem acceptable~\cite{bib:Tofallis2015, bib:Chattopadhyay2016}.

For this reason, the thresholds for the microbe class probability and NMS are selected for each model to minimize first the sMAPE metric, which is defined as
\begin{equation}
    \mathrm{sMAPE} = \frac{100\%}{N}\sum_{i=1}^N\frac{|x_i - \Tilde{x}_i|}{|x_i + \Tilde{x}_i|},
\label{eq:sMape}
\end{equation}
where $N$ is a number of all \textit{countable} samples, $x_i$ is total count of instances present on $i$-th image, and $\Tilde{x}_i$ is predicted number of microbe instances. In the above formula, if $x_{i}=\Tilde{x}_i=0$, then the $i$-th term in the summation is 0 (calculated error is zero, because the number of predicted colonies is correct). Please note that the denominator favors counting errors for samples with a small number of colonies, ensuring that they are equally important as those with more microbe instances. Moreover, the applied symmetrization favors neither underestimated (small recall) nor overestimated (low precision) counts by a model.

If the value of a few lowest sMAPE values (for different thresholds) slightly differs (less than 0.1\%), we compare achieved MAE value calculated using the formula
\begin{equation}
    \mathrm{MAE} = \frac{1}{N}\sum_{i=1}^N|x_i - \Tilde{x}_i|,
\label{eq:MAE}
\end{equation}
and choose such thresholds pair that gives the lowest MAE. Additionally, sum of MAE calculated for every microbe separately, called cumulative MAE, is given as
\begin{equation}
    \mathrm{cMAE} = \sum_{m=1}^K\mathrm{MAE}_m,
\label{eq:cMAE}
\end{equation}
where $K=5$ is a number of all microbe species, and $\mathrm{MAE}_m$ indicates MAE calculated for $m$-th microbial class solely.

\section{Deep learning models evaluation and comparison}

In this section detailed analysis is presented to verify the suitability of the AGAR dataset for building deep learning models for image-based microorganisms recognition. To set a benchmark for the task, the performance of two neural networks architectures for object detection: Faster R-CNN~\cite{bib:Faster2015} and Cascade R-CNN~\cite{bib:Cascade2018}, with four different backbones: ResNet-50\cite{bib:resnet2016}, ResNet-101~\cite{bib:resnet2016}, ResNeXt-101~\cite{bib:resnext2017}, and HRNet~\cite{bib:hrnet2020} is evaluated. The implementation from the MMDetection~\cite{bib:mmdetection2019} toolbox was used. Backbones' weights were initialized using models pre-trained on ImageNet~\cite{bib:imagenet2009}, available in the \textit{torchvision} package of the PyTorch library.

The quantitative results for each subset of AGAR analyzed separately, as well as the results for models trained on the whole dataset, are reported. Each subset was split randomly into approximately 75\% samples constituting the training set, and 25\% for the validation/testing set. Due to the relatively high resolution of all images, they were divided into patches of 512x512~px to avoid scaling them down, which could make it difficult to recognize small colonies. If necessary, zero padding was applied to get a proper aspect ratio. As the goal of this study is to establish a baseline for the microbe detection task, neural network architectures were not adjusted, so the patches were resized to match default input layer size.

Several approaches to data augmentation were tested. During training, Gaussian blur and salt and pepper noise were added, image color space was changed to LAB and HSV, histogram equalization was applied, images were rotated in the $[-45, 45]$ degrees range, and cropped around the annotated bounding boxes, such that there is always a visible microbe object. However, the best results were obtained through splitting the images into patches, and normalizing them using means and standard deviation values per channel as for the COCO dataset~\cite{bib:COCO2014}.

\subsection{Training and validation details}
\label{subsubsec:training_and_validation}
As it was mentioned earlier, patches for validation and testing are prepared from the same photos (1/4 of the given subset), albeit in a different manner: patches for testing are cut out 
from images evenly with some padding applied -- see the blue path in Fig.~\ref{fig:patches}. It turns out that to filter out unwanted bounding boxes during the post-processing stage, two parameters need to be adjusted: detected colony classification probability and NMS threshold. Both of them are fitted to each subset separately. However, to ensure methodological correctness during tests, they are optimized on the subset of training data, and the values which minimize sMAPE is chosen.

The AGAR dataset is relatively large and the top deep learning models used in this studies are of complex structure (e.g. the Cascade R-CNN model with the HRNet has got 956 different layers in total). Therefore, the training process lasts from about 2 days for the simplest (and smallest) Faster R-CNN with ResNet-50 backbone to about 5 days for the biggest one, Cascade R-CNN with the HRNetV2W48 backbone. Calculations were performed on NVIDIA Tesla V100 GPUs. 

During training, the Stochastic Gradient Descent (SGD) method was used for optimizing the network weights, and models were trained over 20 epochs with a batch size of 3-10 patches (depending on the model complexity, i.e. GPU memory consumption), and an adaptive learning rate was initialized with 0.0001. After about 20 epochs loss values saturate, and longer training reduces neither the training nor the validation error.

\subsection{Precision of detection}
\label{subsubsec:detection_precission}
Generally speaking, the precision of detection relates to overlapping between predicted and ground truth bounding boxes, which is measured by the IoU. The prediction is valid if IoU is above given threshold (and a microbe is properly classified), as described in Section~\ref{subsubsec:detection_metrics}. Precision and recall are then calculated (as defined by Eq.~\ref{eq:precision_recall}) for every colony in a subset giving a precision-recall curve.

\begin{figure}[!bt]
\centering
\includegraphics[width=.42\textwidth]{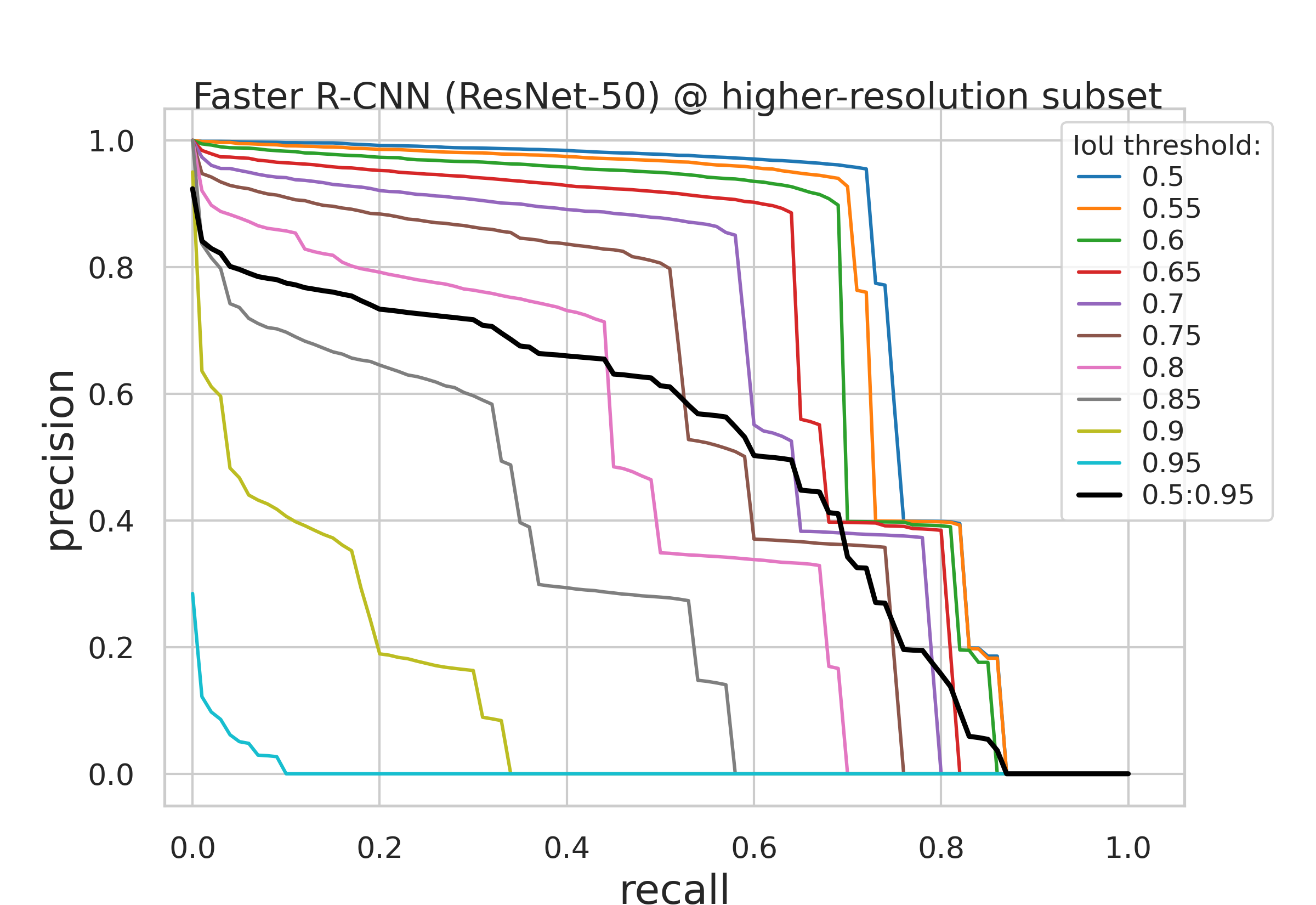}
\includegraphics[width=.42\textwidth]{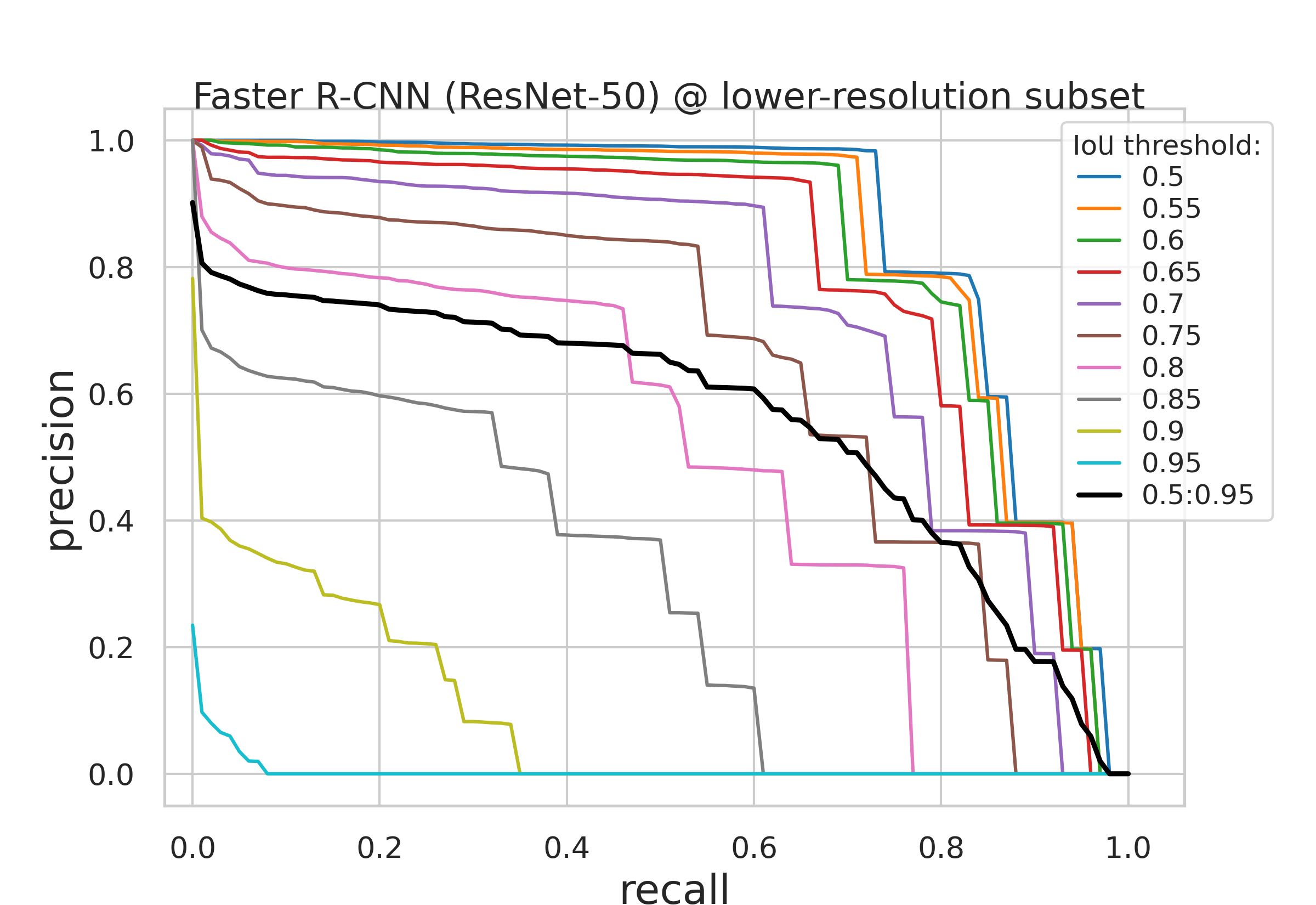}\\
\includegraphics[width=.42\textwidth]{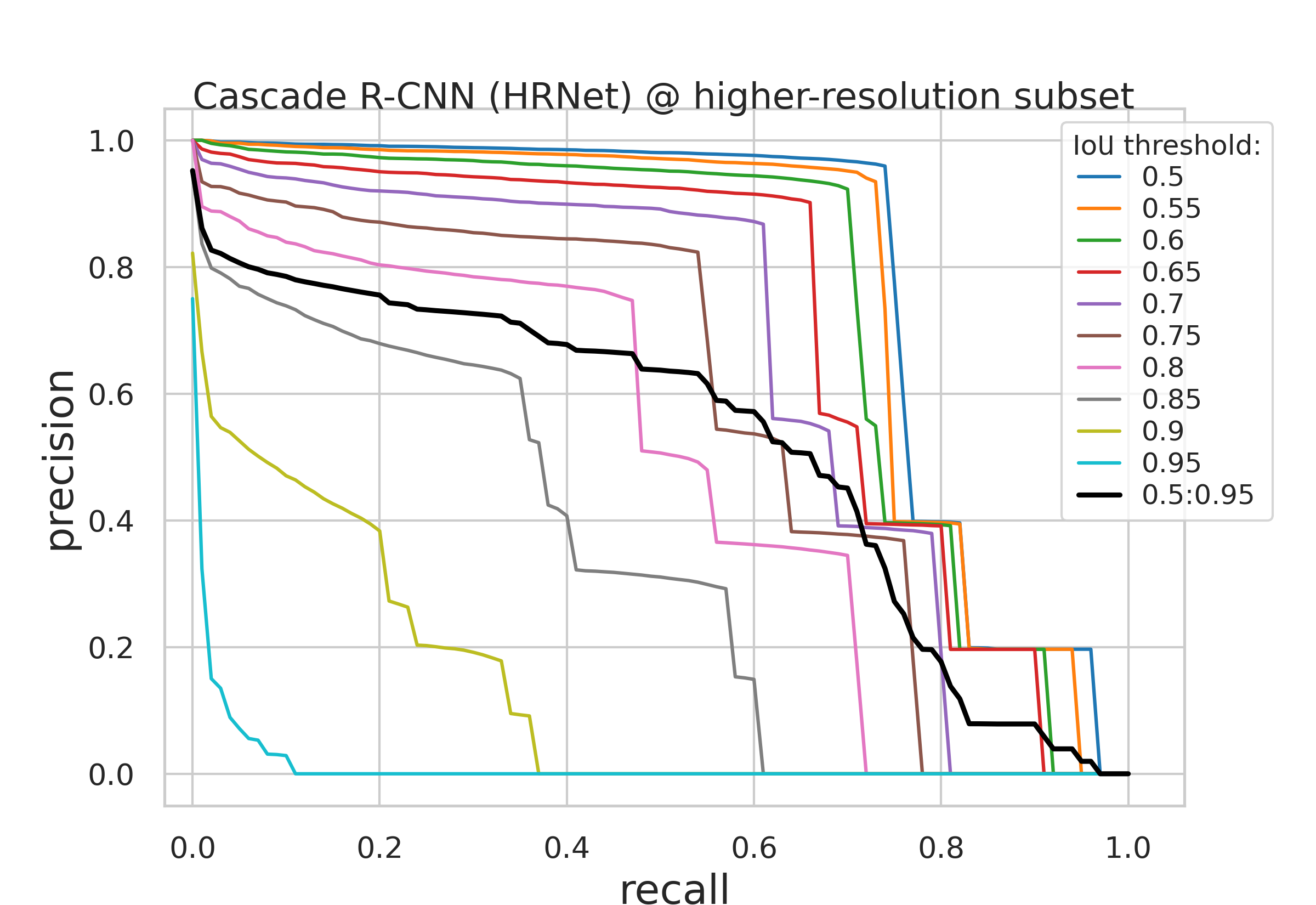}
\includegraphics[width=.42\textwidth]{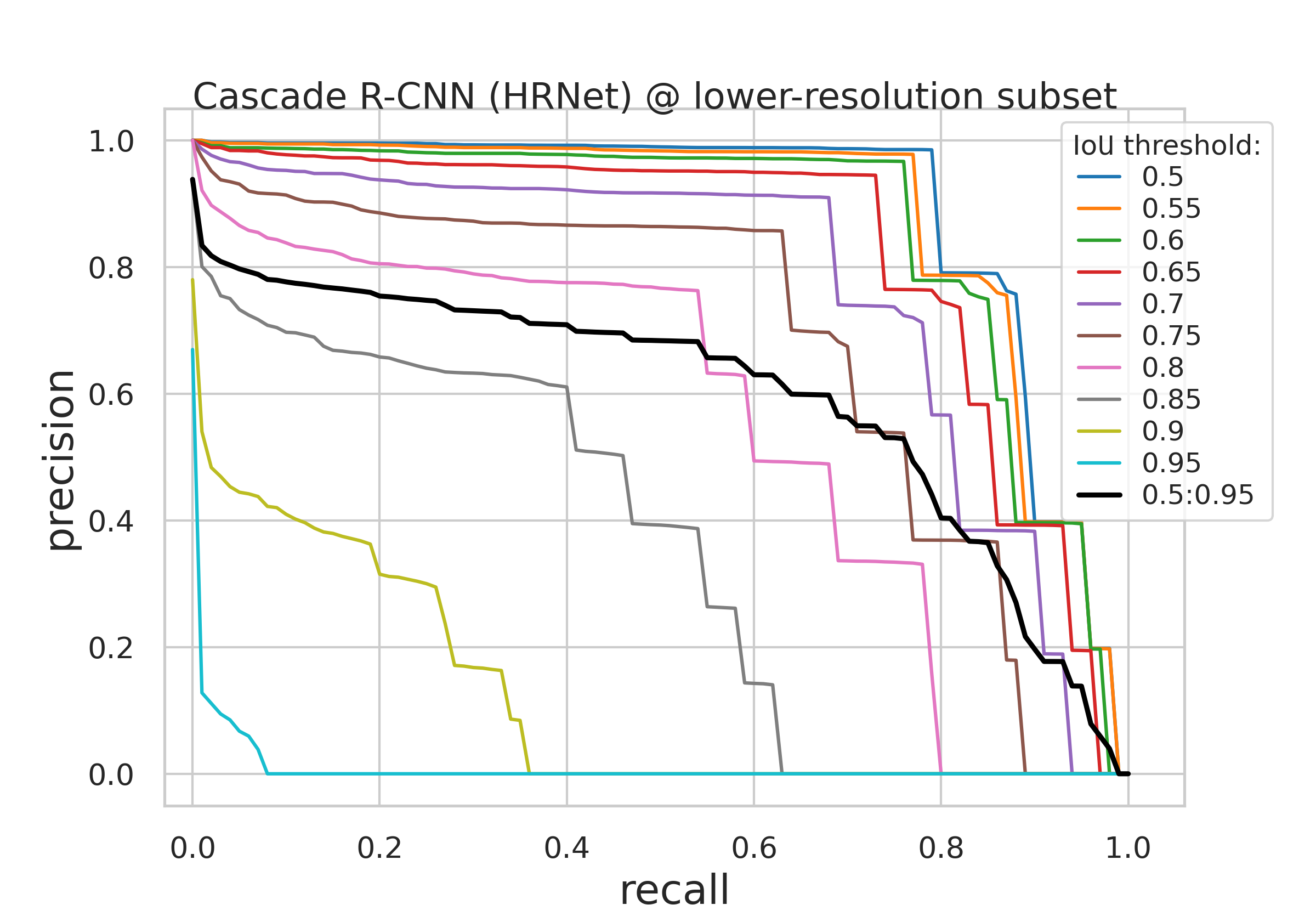}
\caption{\label{fig:precision_recall} The precision-recall curves for different IoU thresholds obtained for the two major subsets of the AGAR dataset using Faster R-CNN with ResNet-50, and Cascade R-CNN with HRNet models.}
\end{figure}
Fig.~\ref{fig:precision_recall} presents the PR curves at different IoU thresholds levels, ranging from $0.5$ to $0.95$, for two selected models on \textit{lower}- and \textit{higher-resolution} data subset separately. The black curve is the average over all thresholds. These curves shows two characteristic features. Firstly, precision competes with recall in a sense that requirement of high detection precision results in low recall, and vice versa. Moreover, the higher IoU threshold (so higher overlapping requirements), the lower precision values. In the ideal case, precision equals one for all the values of recall. Secondly, for high recall values, precision is greater for the \textit{lower-resolution} background category.
\begin{figure}[!hbt]
\centering
\includegraphics[width=.42\textwidth]{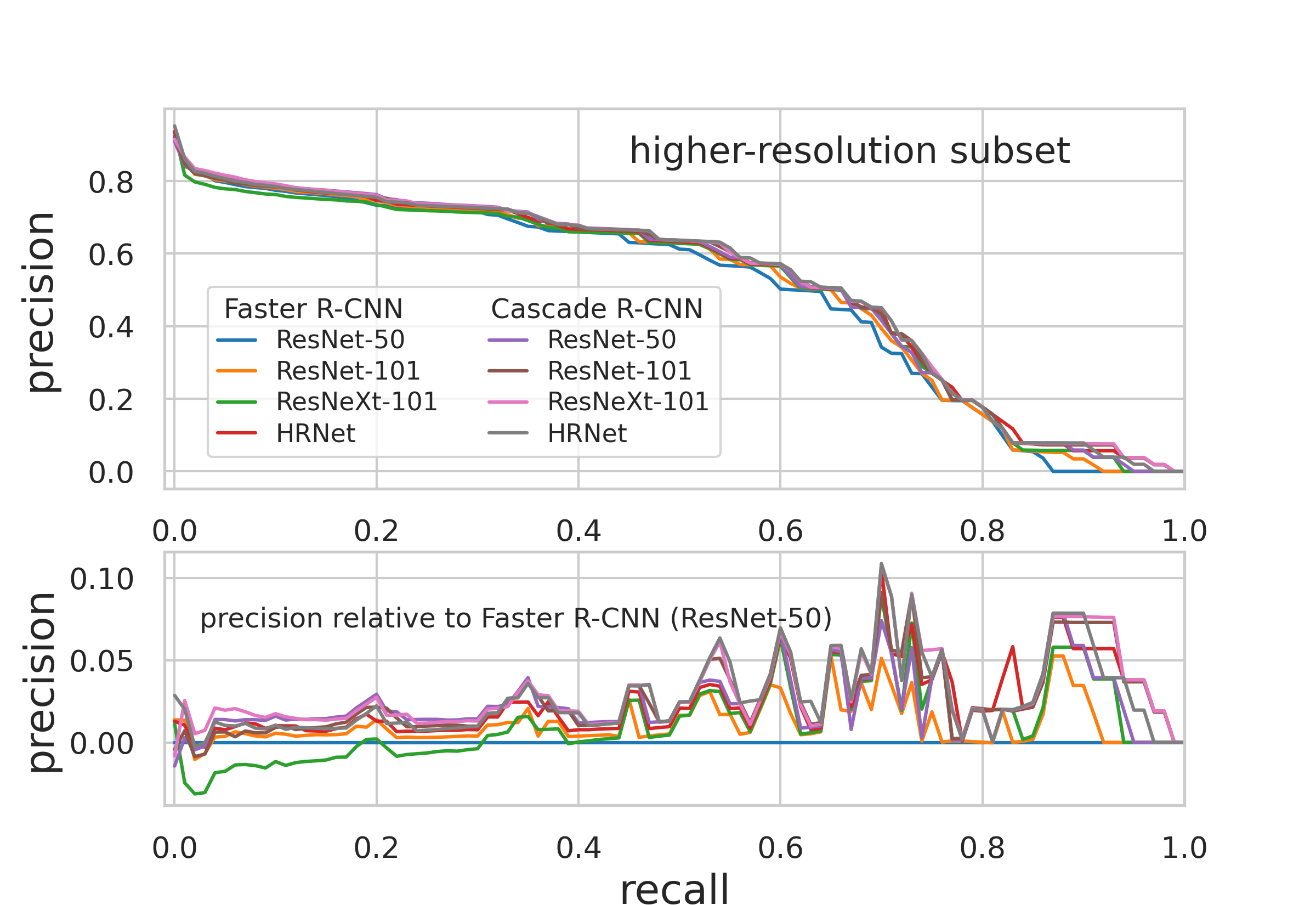}
\includegraphics[width=.42\textwidth]{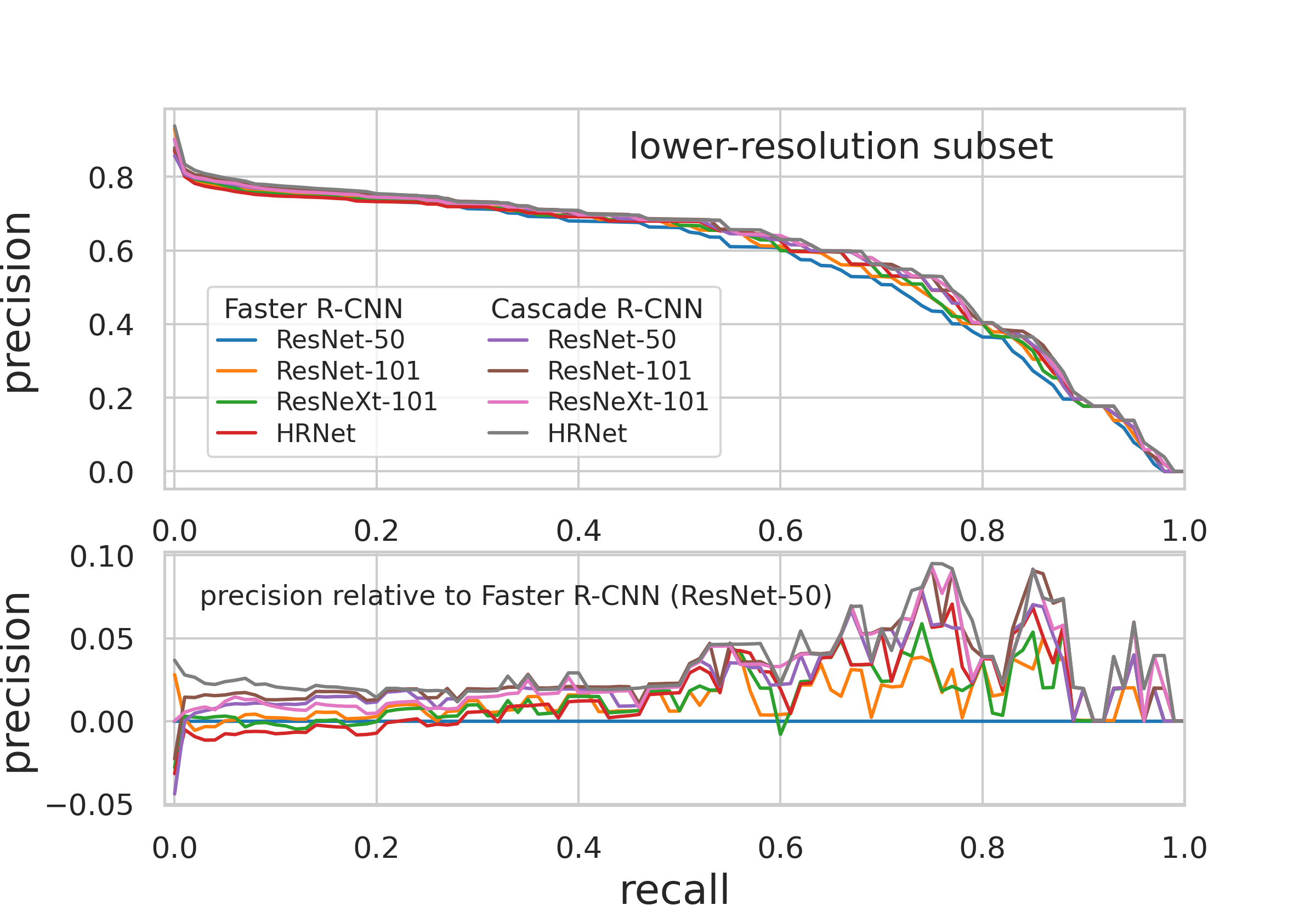}
\caption{\label{fig:precision_recall_models} The precision-recall curves for various models: Faster R-CNN and Cascade R-CNNs with four different backbones, trained on \textit{higher-resolution} and \textit{lower-resolution} subsets. For bottom charts we establish a baseline based on a the Faster R-CNN with ResNet-50, i.e. precision for the all other models are presented relative to this model.}
\end{figure}

The averaged PR curves (over IoU thresholds in the range $[0.5:0.95]$) for different models are presented in Fig.~\ref{fig:precision_recall_models}. Altogether, eight models are considered as the all possible combinations of two detector architectures (Faster R-CNN and Cascade R-CNN) with four different backbones (ResNet-50, ResNet-101, ResNeXt-101, and HRNetV2W48). The smallest model (Faster R-CNN with ResNet-50) is chosen as a baseline and the relative performance of all other models is also show in Fig.~\ref{fig:precision_recall_models}. The best result are obtained for Cascade R-CNN with ResNeXt-101 in the case of \textit{higher-resolution} background category, and with HRNet in the case of \textit{lower-resolution} subset. However, the small difference between the models performance suggests that the models capability (at least in this class of architectures) has been reached, and further models complicating (e.g. by adding more layers in the backbones) may not lead to much better performance.

\begin{table*}[!t]
  \centering
  \caption{\label{tab:ap_model_higher_resolution} Average precision for different models averaged overt different IoU-threshold levels obtained for the \textit{higher-resolution} subset. The Cascade R-CNN model with ResNeXt-101 backbone gives the highest mAP.}
    \footnotesize
    \begin{tabular}{|lcccccccc|}
    \multicolumn{1}{l}{\multirow{2}[0]{*}{IoU}} & \multicolumn{4}{c}{Faster R-CNN}    & \multicolumn{4}{c}{Cascade R-CNN} \\
    \multicolumn{1}{l}{} & ResNet-50 & ResNet-101 & ResNeXt-101 & HRNet & ResNet-50 & ResNet-101 & ResNeXt-101 & \multicolumn{1}{c}{HRNet} \\
    \multicolumn{1}{l}{0.5} & \cellcolor[rgb]{ .482,  .773,  .49}76.7\% & \cellcolor[rgb]{ .451,  .765,  .486}77.9\% & \cellcolor[rgb]{ .431,  .761,  .486}78.5\% & \cellcolor[rgb]{ .396,  .749,  .486}79.9\% & \cellcolor[rgb]{ .416,  .753,  .486}79.2\% & \cellcolor[rgb]{ .396,  .749,  .486}79.9\% & \cellcolor[rgb]{ .388,  .745,  .482}80.1\% & \multicolumn{1}{c}{\cellcolor[rgb]{ .408,  .753,  .486}79.5\%} \\
    \multicolumn{1}{l}{0.55} & \cellcolor[rgb]{ .533,  .788,  .494}74.7\% & \cellcolor[rgb]{ .498,  .776,  .49}76.1\% & \cellcolor[rgb]{ .482,  .773,  .49}76.6\% & \cellcolor[rgb]{ .447,  .765,  .486}78.0\% & \cellcolor[rgb]{ .467,  .769,  .49}77.3\% & \cellcolor[rgb]{ .447,  .765,  .486}78.0\% & \cellcolor[rgb]{ .443,  .761,  .486}78.2\% & \multicolumn{1}{c}{\cellcolor[rgb]{ .455,  .765,  .486}77.8\%} \\
    \multicolumn{1}{l}{0.6} & \cellcolor[rgb]{ .608,  .808,  .498}71.9\% & \cellcolor[rgb]{ .58,  .8,  .494}73.0\% & \cellcolor[rgb]{ .576,  .8,  .494}73.1\% & \cellcolor[rgb]{ .533,  .788,  .494}74.8\% & \cellcolor[rgb]{ .549,  .792,  .494}74.2\% & \cellcolor[rgb]{ .529,  .788,  .494}74.9\% & \cellcolor[rgb]{ .522,  .784,  .49}75.2\% & \multicolumn{1}{c}{\cellcolor[rgb]{ .537,  .788,  .494}74.6\%} \\
    \multicolumn{1}{l}{0.65} & \cellcolor[rgb]{ .729,  .843,  .502}67.3\% & \cellcolor[rgb]{ .694,  .835,  .502}68.6\% & \cellcolor[rgb]{ .702,  .835,  .502}68.4\% & \cellcolor[rgb]{ .647,  .824,  .498}70.3\% & \cellcolor[rgb]{ .651,  .824,  .498}70.4\% & \cellcolor[rgb]{ .624,  .816,  .498}71.3\% & \cellcolor[rgb]{ .62,  .812,  .498}71.5\% & \multicolumn{1}{c}{\cellcolor[rgb]{ .643,  .82,  .498}70.7\%} \\
    \multicolumn{1}{l}{0.7} & \cellcolor[rgb]{ .871,  .886,  .514}61.9\% & \cellcolor[rgb]{ .839,  .878,  .51}63.1\% & \cellcolor[rgb]{ .839,  .878,  .51}63.1\% & \cellcolor[rgb]{ .816,  .871,  .51}64.0\% & \cellcolor[rgb]{ .816,  .871,  .51}64.2\% & \cellcolor[rgb]{ .812,  .871,  .51}64.2\% & \cellcolor[rgb]{ .808,  .867,  .51}64.5\% & \multicolumn{1}{c}{\cellcolor[rgb]{ .812,  .871,  .51}64.3\%} \\
    \multicolumn{1}{l}{0.75} & \cellcolor[rgb]{ .996,  .902,  .514}54.8\% & \cellcolor[rgb]{ .996,  .91,  .514}55.9\% & \cellcolor[rgb]{ .996,  .91,  .514}55.7\% & \cellcolor[rgb]{ .996,  .922,  .518}57.1\% & \cellcolor[rgb]{ .996,  .918,  .514}57.0\% & \cellcolor[rgb]{ .996,  .918,  .514}56.9\% & \cellcolor[rgb]{ .988,  .922,  .518}57.6\% & \multicolumn{1}{c}{\cellcolor[rgb]{ .996,  .922,  .518}57.3\%} \\
    \multicolumn{1}{l}{0.8} & \cellcolor[rgb]{ .992,  .804,  .494}44.4\% & \cellcolor[rgb]{ .992,  .816,  .494}45.6\% & \cellcolor[rgb]{ .992,  .82,  .498}46.0\% & \cellcolor[rgb]{ .992,  .831,  .498}47.4\% & \cellcolor[rgb]{ .992,  .831,  .498}47.3\% & \cellcolor[rgb]{ .992,  .827,  .498}47.2\% & \cellcolor[rgb]{ .992,  .839,  .502}48.2\% & \multicolumn{1}{c}{\cellcolor[rgb]{ .992,  .835,  .498}47.9\%} \\
    \multicolumn{1}{l}{0.85} & \cellcolor[rgb]{ .984,  .671,  .467}29.4\% & \cellcolor[rgb]{ .984,  .682,  .471}30.8\% & \cellcolor[rgb]{ .984,  .682,  .471}30.9\% & \cellcolor[rgb]{ .984,  .694,  .471}32.2\% & \cellcolor[rgb]{ .984,  .698,  .475}32.7\% & \cellcolor[rgb]{ .984,  .694,  .475}32.5\% & \cellcolor[rgb]{ .988,  .702,  .475}33.2\% & \multicolumn{1}{c}{\cellcolor[rgb]{ .988,  .702,  .475}33.2\%} \\
    \multicolumn{1}{l}{0.9} & \cellcolor[rgb]{ .976,  .502,  .435}11.1\% & \cellcolor[rgb]{ .976,  .51,  .435}11.9\% & \cellcolor[rgb]{ .976,  .51,  .435}11.8\% & \cellcolor[rgb]{ .976,  .518,  .439}12.5\% & \cellcolor[rgb]{ .976,  .518,  .439}12.8\% & \cellcolor[rgb]{ .976,  .518,  .439}13.0\% & \cellcolor[rgb]{ .976,  .518,  .439}12.9\% & \multicolumn{1}{c}{\cellcolor[rgb]{ .976,  .522,  .439}13.2\%} \\
    \multicolumn{1}{l}{0.95} & \cellcolor[rgb]{ .973,  .412,  .42}0.8\% & \cellcolor[rgb]{ .973,  .416,  .42}1.4\% & \cellcolor[rgb]{ .973,  .412,  .42}1.0\% & \cellcolor[rgb]{ .973,  .416,  .42}1.3\% & \cellcolor[rgb]{ .973,  .412,  .42}1.1\% & \cellcolor[rgb]{ .973,  .416,  .42}1.6\% & \cellcolor[rgb]{ .973,  .416,  .42}1.5\% & \multicolumn{1}{c}{\cellcolor[rgb]{ .973,  .416,  .42}1.7\%} \\
    \midrule
    0.5:0.95 & \cellcolor[rgb]{ 1,  .937,  .612}49.3\% & \cellcolor[rgb]{ .776,  .867,  .565}50.4\% & \cellcolor[rgb]{ .757,  .863,  .561}50.5\% & \cellcolor[rgb]{ .498,  .78,  .506}51.8\% & \cellcolor[rgb]{ 1,  .937,  .612}51.6\% & \cellcolor[rgb]{ .69,  .839,  .549}51.9\% & \cellcolor[rgb]{ .388,  .745,  .482}\textbf{52.3\%} & \cellcolor[rgb]{ .627,  .824,  .533}52.0\% \\
    \bottomrule
    \end{tabular}%
\end{table*}%

\begin{table*}[!t]
  \centering
  \caption{\label{tab:ap_model_lower_resolution} Average precision for different models averaged overt different IoU-threshold levels obtained for the \textit{lower-resolution} subset. The Cascade R-CNN model with HRNet backbone gives the highest mAP.}
    \footnotesize
    \begin{tabular}{|lcccccccc|}
    \multicolumn{1}{l}{\multirow{2}[0]{*}{IoU}} & \multicolumn{4}{c}{Faster R-CNN}    & \multicolumn{4}{c}{Cascade R-CNN} \\
    \multicolumn{1}{l}{} & ResNet-50 & ResNet-101 & ResNeXt-101 & HRNet & ResNet-50 & ResNet-101 & ResNeXt-101 & \multicolumn{1}{c}{HRNet} \\
    \multicolumn{1}{l}{0.5} & \cellcolor[rgb]{ .467,  .769,  .49}86.5\% & \cellcolor[rgb]{ .431,  .761,  .486}87.6\% & \cellcolor[rgb]{ .439,  .761,  .486}87.4\% & \cellcolor[rgb]{ .42,  .757,  .486}88.2\% & \cellcolor[rgb]{ .408,  .753,  .486}88.6\% & \cellcolor[rgb]{ .388,  .745,  .482}89.2\% & \cellcolor[rgb]{ .404,  .753,  .486}88.7\% & \multicolumn{1}{c}{\cellcolor[rgb]{ .396,  .749,  .486}89.1\%} \\
    \multicolumn{1}{l}{0.55} & \cellcolor[rgb]{ .506,  .78,  .49}85.0\% & \cellcolor[rgb]{ .471,  .769,  .49}86.3\% & \cellcolor[rgb]{ .455,  .765,  .486}86.7\% & \cellcolor[rgb]{ .447,  .765,  .486}87.1\% & \cellcolor[rgb]{ .439,  .761,  .486}87.4\% & \cellcolor[rgb]{ .427,  .757,  .486}87.9\% & \cellcolor[rgb]{ .435,  .761,  .486}87.6\% & \multicolumn{1}{c}{\cellcolor[rgb]{ .427,  .757,  .486}87.9\%} \\
    \multicolumn{1}{l}{0.6} & \cellcolor[rgb]{ .553,  .796,  .494}83.2\% & \cellcolor[rgb]{ .522,  .784,  .49}84.4\% & \cellcolor[rgb]{ .522,  .784,  .49}84.3\% & \cellcolor[rgb]{ .51,  .78,  .49}84.8\% & \cellcolor[rgb]{ .498,  .776,  .49}85.5\% & \cellcolor[rgb]{ .478,  .773,  .49}86.1\% & \cellcolor[rgb]{ .475,  .773,  .49}86.2\% & \multicolumn{1}{c}{\cellcolor[rgb]{ .478,  .773,  .49}86.1\%} \\
    \multicolumn{1}{l}{0.65} & \cellcolor[rgb]{ .655,  .824,  .498}79.5\% & \cellcolor[rgb]{ .616,  .812,  .498}80.8\% & \cellcolor[rgb]{ .62,  .812,  .498}80.7\% & \cellcolor[rgb]{ .612,  .812,  .498}81.1\% & \cellcolor[rgb]{ .592,  .804,  .494}82.1\% & \cellcolor[rgb]{ .576,  .8,  .494}82.6\% & \cellcolor[rgb]{ .592,  .804,  .494}82.1\% & \multicolumn{1}{c}{\cellcolor[rgb]{ .576,  .8,  .494}82.6\%} \\
    \multicolumn{1}{l}{0.7} & \cellcolor[rgb]{ .824,  .871,  .51}73.2\% & \cellcolor[rgb]{ .773,  .859,  .506}75.0\% & \cellcolor[rgb]{ .773,  .855,  .506}75.2\% & \cellcolor[rgb]{ .769,  .855,  .506}75.2\% & \cellcolor[rgb]{ .749,  .851,  .506}76.6\% & \cellcolor[rgb]{ .737,  .847,  .506}77.1\% & \cellcolor[rgb]{ .745,  .851,  .506}76.7\% & \multicolumn{1}{c}{\cellcolor[rgb]{ .753,  .851,  .506}76.5\%} \\
    \multicolumn{1}{l}{0.75} & \cellcolor[rgb]{ .996,  .898,  .51}63.6\% & \cellcolor[rgb]{ .996,  .91,  .514}65.4\% & \cellcolor[rgb]{ .996,  .918,  .514}66.1\% & \cellcolor[rgb]{ .996,  .914,  .514}65.8\% & \cellcolor[rgb]{ .996,  .914,  .514}67.1\% & \cellcolor[rgb]{ .988,  .918,  .518}68.3\% & \cellcolor[rgb]{ .996,  .918,  .514}67.5\% & \multicolumn{1}{c}{\cellcolor[rgb]{ .996,  .922,  .518}68.0\%} \\
    \multicolumn{1}{l}{0.8} & \cellcolor[rgb]{ .992,  .788,  .49}49.6\% & \cellcolor[rgb]{ .992,  .8,  .494}51.3\% & \cellcolor[rgb]{ .992,  .804,  .494}51.7\% & \cellcolor[rgb]{ .992,  .808,  .494}52.2\% & \cellcolor[rgb]{ .992,  .812,  .494}53.4\% & \cellcolor[rgb]{ .992,  .816,  .494}54.0\% & \cellcolor[rgb]{ .992,  .812,  .494}53.7\% & \multicolumn{1}{c}{\cellcolor[rgb]{ .992,  .824,  .498}55.0\%} \\
    \multicolumn{1}{l}{0.85} & \cellcolor[rgb]{ .984,  .631,  .459}29.5\% & \cellcolor[rgb]{ .984,  .643,  .463}30.9\% & \cellcolor[rgb]{ .984,  .643,  .463}30.9\% & \cellcolor[rgb]{ .984,  .647,  .463}31.1\% & \cellcolor[rgb]{ .984,  .647,  .463}32.1\% & \cellcolor[rgb]{ .984,  .655,  .467}32.9\% & \cellcolor[rgb]{ .984,  .655,  .463}32.7\% & \multicolumn{1}{c}{\cellcolor[rgb]{ .984,  .671,  .467}35.1\%} \\
    \multicolumn{1}{l}{0.9} & \cellcolor[rgb]{ .973,  .475,  .431}9.2\% & \cellcolor[rgb]{ .973,  .482,  .431}9.8\% & \cellcolor[rgb]{ .976,  .482,  .431}10.0\% & \cellcolor[rgb]{ .976,  .482,  .431}10.0\% & \cellcolor[rgb]{ .976,  .486,  .431}10.7\% & \cellcolor[rgb]{ .976,  .49,  .431}11.2\% & \cellcolor[rgb]{ .976,  .49,  .431}11.0\% & \multicolumn{1}{c}{\cellcolor[rgb]{ .976,  .498,  .435}12.2\%} \\
    \multicolumn{1}{l}{0.95} & \cellcolor[rgb]{ .973,  .412,  .42}0.6\% & \cellcolor[rgb]{ .973,  .412,  .42}0.9\% & \cellcolor[rgb]{ .973,  .412,  .42}0.7\% & \cellcolor[rgb]{ .973,  .412,  .42}0.7\% & \cellcolor[rgb]{ .973,  .412,  .42}0.7\% & \cellcolor[rgb]{ .973,  .412,  .42}0.9\% & \cellcolor[rgb]{ .973,  .412,  .42}0.9\% & \multicolumn{1}{c}{\cellcolor[rgb]{ .973,  .416,  .42}1.2\%} \\
    \midrule
    0.5:0.95 & \cellcolor[rgb]{ 1,  .937,  .612}56.0\% & \cellcolor[rgb]{ .773,  .867,  .565}57.3\% & \cellcolor[rgb]{ .749,  .859,  .561}57.4\% & \cellcolor[rgb]{ .706,  .847,  .553}57.6\% & \cellcolor[rgb]{ 1,  .937,  .612}58.4\% & \cellcolor[rgb]{ .624,  .82,  .533}59.0\% & \cellcolor[rgb]{ .804,  .878,  .573}58.7\% & \cellcolor[rgb]{ .388,  .745,  .482}\textbf{59.4\%} \\
    \bottomrule
    \end{tabular}%
\end{table*}%

The average precision values at different IoU thresholds are presented in Table~\ref{tab:ap_model_higher_resolution} for the \textit{higher-resolution} subset, and in Table~\ref{tab:ap_model_lower_resolution} for the \textit{lower-resolution} one. The last row is mean AP (mAP) over the whole range of thresholds, i.e. $[0.5:0.95]$. In general, Cascade R-CNN performs slightly better on the AGAR dataset than Faster R-CNN. Although increasing the complexity of a backbones gives slightly better results, most of them are quite close to each other.

The results presented in Figures~\ref{fig:precision_recall}-\ref{fig:precision_recall_models} and Tables~\ref{tab:ap_model_higher_resolution}-\ref{tab:ap_model_lower_resolution} are averaged over the all microbe classes. Per-microbe precision is shown in Tab.~\ref{tab:ap_class_faster} for the baseline model compared to Cascade R-CNN with the HRNet backbone, on the \textit{higher-resolution} subset. In general, detector performs better for microbes that form smaller colonies, i.e. \textit{S. aureus} and \textit{C. albicans}. The lower precision values for bigger colonies are mainly due to their tendency to aggregate and overlap, especially in higher populated samples. Such overlapped colonies make it harder to detect individual colonies. Moreover, larger colonies tend to have blurred edges (with the lower contrast relative to agar substrate), which affects IoU.

\begin{table*}[!t]
  \footnotesize
  \centering
  \caption{\label{tab:ap_class_faster} Average precision calculated for each class of microbes separately at different IoU-threshold levels for (left) Faster R-CNN (ResNet-50), and (right) Cascade R-CNN (HRNet) on \textit{higher-resolution} subset. We get overall better performance for smaller colonies.}
    \begin{tabular}{|lccccc|ccccc|}
    \multicolumn{1}{p{5em}}{} & \multicolumn{5}{c|}{AP per class: Faster R-CNN (ResNet-50)} & \multicolumn{5}{c}{AP per class: Cascade R-CNN (HRNet)} \\
    \multicolumn{1}{p{5em}}{IoU} & \textcolor[rgb]{ .09,  .169,  .302}{\textit{S. aureus}} & \textcolor[rgb]{ .09,  .169,  .302}{\textit{B. subtilis}} & \textcolor[rgb]{ .09,  .169,  .302}{\textit{P. aeruginosa}} & \textcolor[rgb]{ .09,  .169,  .302}{\textit{E. coli}} & \textcolor[rgb]{ .09,  .169,  .302}{\textit{C. albicans}} & \textcolor[rgb]{ .09,  .169,  .302}{\textit{S. aureus}} & \textcolor[rgb]{ .09,  .169,  .302}{\textit{B. subtilis}} & \textcolor[rgb]{ .09,  .169,  .302}{\textit{P. aeruginosa}} & \textcolor[rgb]{ .09,  .169,  .302}{\textit{E. coli}} & \multicolumn{1}{c}{\textcolor[rgb]{ .09,  .169,  .302}{\textit{C. albicans}}} \\
    \multicolumn{1}{l}{0.5} & \cellcolor[rgb]{ .388,  .745,  .482}86.0\% & \cellcolor[rgb]{ .675,  .827,  .502}72.4\% & \cellcolor[rgb]{ .729,  .843,  .502}69.9\% & \cellcolor[rgb]{ .663,  .824,  .498}73.1\% & \cellcolor[rgb]{ .475,  .773,  .49}82.0\% & \cellcolor[rgb]{ .388,  .745,  .482}95.8\% & \cellcolor[rgb]{ .761,  .855,  .506}73.6\% & \cellcolor[rgb]{ .792,  .863,  .506}71.7\% & \cellcolor[rgb]{ .745,  .851,  .506}74.6\% & \multicolumn{1}{c}{\cellcolor[rgb]{ .62,  .812,  .498}82.1\%} \\
    \multicolumn{1}{l}{0.55} & \cellcolor[rgb]{ .392,  .749,  .486}85.9\% & \cellcolor[rgb]{ .737,  .847,  .506}69.5\% & \cellcolor[rgb]{ .78,  .859,  .506}67.4\% & \cellcolor[rgb]{ .749,  .851,  .506}68.9\% & \cellcolor[rgb]{ .475,  .773,  .49}82.0\% & \cellcolor[rgb]{ .424,  .757,  .486}93.8\% & \cellcolor[rgb]{ .808,  .867,  .51}70.8\% & \cellcolor[rgb]{ .812,  .867,  .51}70.7\% & \cellcolor[rgb]{ .796,  .863,  .506}71.5\% & \multicolumn{1}{c}{\cellcolor[rgb]{ .62,  .812,  .498}82.0\%} \\
    \multicolumn{1}{l}{0.6} & \cellcolor[rgb]{ .416,  .753,  .486}84.8\% & \cellcolor[rgb]{ .839,  .878,  .51}64.6\% & \cellcolor[rgb]{ .82,  .871,  .51}65.6\% & \cellcolor[rgb]{ .855,  .882,  .51}64.0\% & \cellcolor[rgb]{ .498,  .776,  .49}80.8\% & \cellcolor[rgb]{ .475,  .773,  .49}90.8\% & \cellcolor[rgb]{ .894,  .894,  .514}65.7\% & \cellcolor[rgb]{ .839,  .878,  .51}68.9\% & \cellcolor[rgb]{ .878,  .886,  .514}66.6\% & \multicolumn{1}{c}{\cellcolor[rgb]{ .639,  .82,  .498}80.9\%} \\
    \multicolumn{1}{l}{0.65} & \cellcolor[rgb]{ .498,  .776,  .49}80.9\% & \cellcolor[rgb]{ 1,  .922,  .518}56.9\% & \cellcolor[rgb]{ .886,  .89,  .514}62.4\% & \cellcolor[rgb]{ .996,  .918,  .514}56.6\% & \cellcolor[rgb]{ .525,  .784,  .49}79.6\% & \cellcolor[rgb]{ .49,  .776,  .49}89.7\% & \cellcolor[rgb]{ .996,  .918,  .514}59.1\% & \cellcolor[rgb]{ .898,  .894,  .514}65.5\% & \cellcolor[rgb]{ 1,  .922,  .518}59.3\% & \multicolumn{1}{c}{\cellcolor[rgb]{ .659,  .824,  .498}79.8\%} \\
    \multicolumn{1}{l}{0.7} & \cellcolor[rgb]{ .561,  .796,  .494}77.9\% & \cellcolor[rgb]{ .996,  .847,  .502}48.9\% & \cellcolor[rgb]{ .98,  .918,  .518}57.9\% & \cellcolor[rgb]{ .992,  .839,  .502}47.9\% & \cellcolor[rgb]{ .576,  .8,  .494}77.1\% & \cellcolor[rgb]{ .667,  .827,  .502}79.4\% & \cellcolor[rgb]{ .992,  .847,  .502}50.9\% & \cellcolor[rgb]{ .965,  .914,  .518}61.5\% & \cellcolor[rgb]{ .996,  .851,  .502}51.2\% & \multicolumn{1}{c}{\cellcolor[rgb]{ .682,  .831,  .502}78.4\%} \\
    \multicolumn{1}{l}{0.75} & \cellcolor[rgb]{ .706,  .839,  .502}71.1\% & \cellcolor[rgb]{ .988,  .765,  .486}39.8\% & \cellcolor[rgb]{ .996,  .867,  .506}50.9\% & \cellcolor[rgb]{ .988,  .757,  .486}39.0\% & \cellcolor[rgb]{ .659,  .824,  .498}73.2\% & \cellcolor[rgb]{ .749,  .851,  .506}74.4\% & \cellcolor[rgb]{ .988,  .765,  .486}41.6\% & \cellcolor[rgb]{ .996,  .871,  .506}53.8\% & \cellcolor[rgb]{ .988,  .769,  .486}41.7\% & \multicolumn{1}{c}{\cellcolor[rgb]{ .741,  .847,  .506}74.8\%} \\
    \multicolumn{1}{l}{0.8} & \cellcolor[rgb]{ .973,  .914,  .518}58.4\% & \cellcolor[rgb]{ .984,  .678,  .471}30.0\% & \cellcolor[rgb]{ .988,  .749,  .482}38.0\% & \cellcolor[rgb]{ .984,  .678,  .471}30.2\% & \cellcolor[rgb]{ .824,  .871,  .51}65.5\% & \cellcolor[rgb]{ .914,  .898,  .514}64.5\% & \cellcolor[rgb]{ .984,  .675,  .471}31.2\% & \cellcolor[rgb]{ .992,  .776,  .486}42.7\% & \cellcolor[rgb]{ .984,  .69,  .471}33.1\% & \multicolumn{1}{c}{\cellcolor[rgb]{ .855,  .882,  .51}68.1\%} \\
    \multicolumn{1}{l}{0.85} & \cellcolor[rgb]{ .988,  .741,  .482}37.4\% & \cellcolor[rgb]{ .98,  .565,  .447}17.5\% & \cellcolor[rgb]{ .98,  .616,  .459}23.3\% & \cellcolor[rgb]{ .98,  .58,  .451}19.4\% & \cellcolor[rgb]{ .996,  .851,  .502}49.4\% & \cellcolor[rgb]{ .992,  .808,  .494}46.5\% & \cellcolor[rgb]{ .98,  .569,  .447}19.2\% & \cellcolor[rgb]{ .98,  .624,  .459}25.5\% & \cellcolor[rgb]{ .98,  .596,  .455}22.3\% & \multicolumn{1}{c}{\cellcolor[rgb]{ .996,  .859,  .506}52.4\%} \\
    \multicolumn{1}{l}{0.9} & \cellcolor[rgb]{ .976,  .541,  .443}14.8\% & \cellcolor[rgb]{ .973,  .463,  .427}6.3\% & \cellcolor[rgb]{ .973,  .478,  .431}7.9\% & \cellcolor[rgb]{ .973,  .475,  .431}7.4\% & \cellcolor[rgb]{ .98,  .58,  .451}19.2\% & \cellcolor[rgb]{ .98,  .561,  .447}18.0\% & \cellcolor[rgb]{ .973,  .471,  .427}7.6\% & \cellcolor[rgb]{ .973,  .475,  .431}8.3\% & \cellcolor[rgb]{ .976,  .49,  .435}10.2\% & \multicolumn{1}{c}{\cellcolor[rgb]{ .98,  .596,  .455}22.1\%} \\
    \multicolumn{1}{l}{0.95} & \cellcolor[rgb]{ .973,  .412,  .42}0.6\% & \cellcolor[rgb]{ .973,  .412,  .42}0.4\% & \cellcolor[rgb]{ .973,  .412,  .42}0.7\% & \cellcolor[rgb]{ .973,  .412,  .42}0.5\% & \cellcolor[rgb]{ .973,  .424,  .42}1.8\% & \cellcolor[rgb]{ .973,  .427,  .42}2.9\% & \cellcolor[rgb]{ .973,  .416,  .42}1.3\% & \cellcolor[rgb]{ .973,  .412,  .42}0.8\% & \cellcolor[rgb]{ .973,  .412,  .42}0.8\% & \multicolumn{1}{c}{\cellcolor[rgb]{ .973,  .427,  .42}2.7\%} \\
    \midrule
    0.5:0.95 & \cellcolor[rgb]{ .941,  .906,  .518}59.8\% & \cellcolor[rgb]{ .988,  .773,  .486}40.6\% & \cellcolor[rgb]{ .992,  .808,  .494}44.4\% & \cellcolor[rgb]{ .988,  .773,  .486}40.7\% & \cellcolor[rgb]{ .914,  .898,  .514}\textbf{61.1\%} & \cellcolor[rgb]{ .898,  .894,  .514}\textbf{65.6\%} & \cellcolor[rgb]{ .988,  .769,  .486}42.1\% & \cellcolor[rgb]{ .992,  .812,  .494}46.9\% & \cellcolor[rgb]{ .992,  .78,  .49}43.1\% & \cellcolor[rgb]{ .949,  .91,  .518}62.3\% \\
    \bottomrule
    \end{tabular}%
\end{table*}%

\subsection{Microbe counting results}
\label{subsubsec:counting_results}

When it comes to counting, the recognition of microbes is crucial, but the exactness of their localization (i.e. true and predicted bounding boxes overlapping) is of secondary importance. In the post processing stage, the sMAPE and MAE metrics are used to tune the probability and NMS thresholds on the AGAR training set, as described in Section~\ref{subsubsec:counting_metrics}. The thresholds are used when the predictions for individual patches are merged into the whole test image prediction (to get rid of the excess of bounding boxes, e.g. appearing twice through the edges of neighbouring patches).

\begin{figure}[!h]
\centering
\includegraphics[width=.45\textwidth]{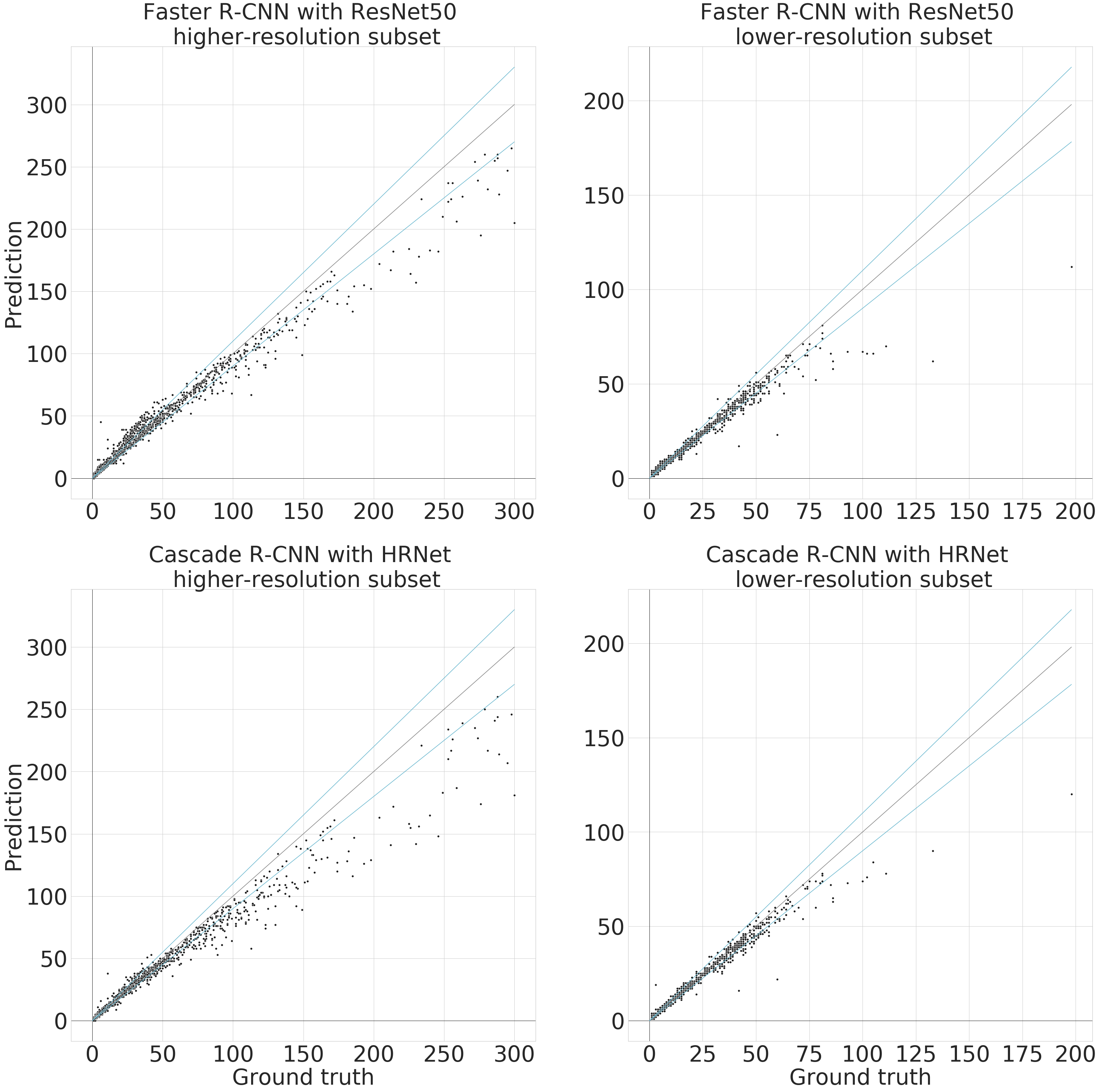}
\hspace{0.5cm}
\includegraphics[width=.45\textwidth]{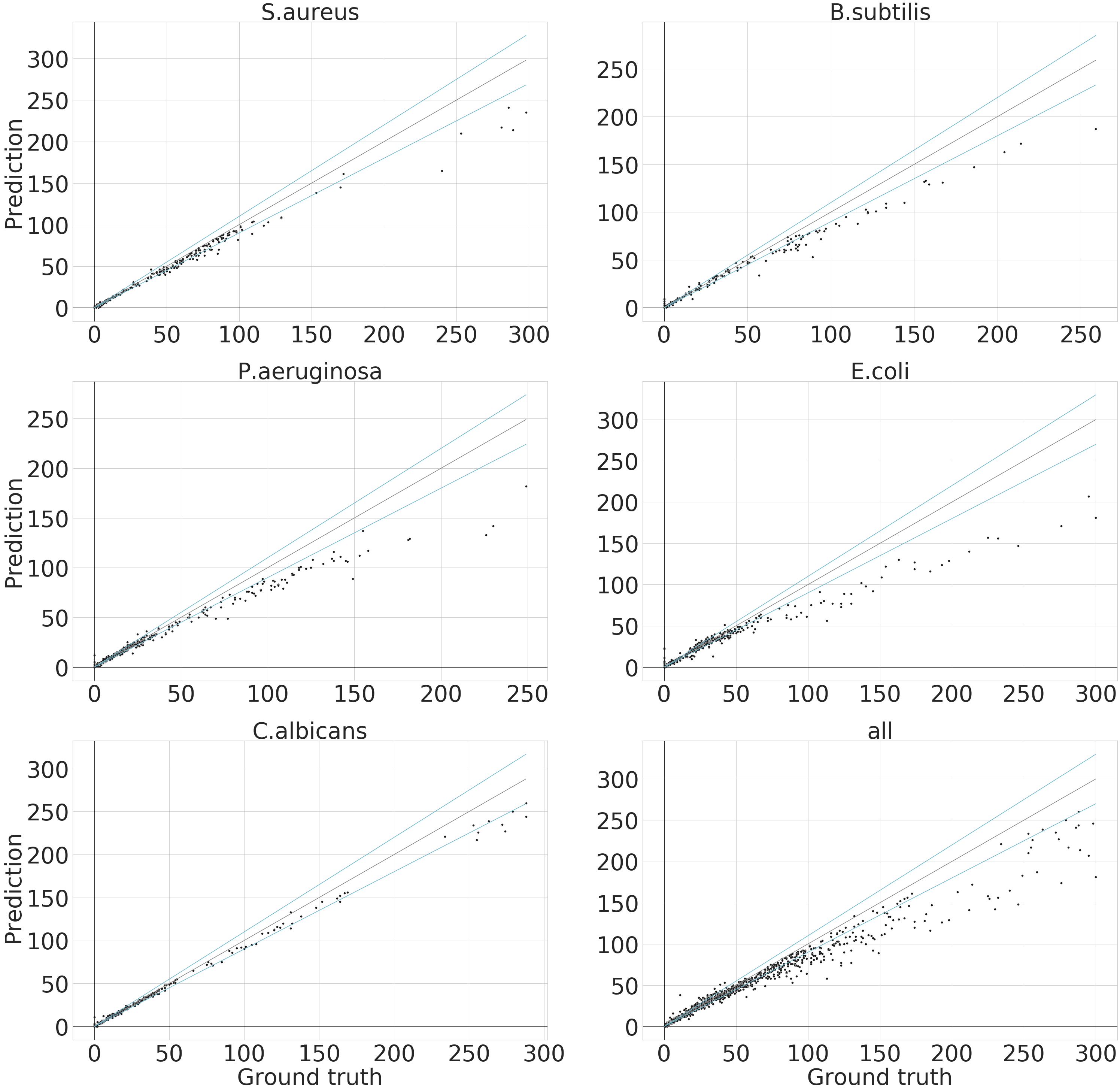}
\caption{\label{fig:gt_vs_pred_4_types}The performance of microbial colony counting on: (left) two main background categories of the AGAR dataset using Faster R-CNN and Cascade R-CNN with two backbones ResNet-50 and HRNet, and (right) results achieved using Cascade R-CNN with HRNet divided into 5 microbial species.}
\end{figure}

The performance of Faster R-CNN and Cascade R-CNN models with ResNet-50 and HRNet backbones for microbial counting is presented in Fig.~\ref{fig:gt_vs_pred_4_types}(left) for both major background categories of the AGAR dataset, and Fig.~\ref{fig:gt_vs_pred_4_types}(right) for all five bacteria species separately. The straight black line represents the identity function (meaning perfect prediction), while cyan lines indicate 10\% error to highlight the acceptable range. Starting from a around 50 colonies, the detectors clearly underestimate the their exact number. This is largely due to the nature of sMAPE, which give more weight to errors coming from less populated samples. Moreover, the AGAR dataset is dominated by the plates with less than 50 colonies. Therefore, the probability and NMS thresholds for post processing are tuned to favour these samples.

Microbe counting metrics for various models for the \textit{higher}- and \textit{lower-resolution} subsets are presented in Tables~\ref{tab:mape_bright}-\ref{tab:mape_lower-resolution}. While the best results are obtained with Cascade R-CNN with HRNet, it is worth noting that the most lightweight model (Faster R-CNN with ResNet-50) is not much worse. In general, the errors for the \textit{lower-resolution} subset are smaller, which is likely caused by the presence of more difficult samples in the \textit{higher-resolution} subset. While the \textit{lower-resolution} subset is more homogeneous, therefore it is simpler. Moreover, the counting performance is better for the smallest microbial colonies (\textit{C. albicans} and \textit{S. aureus}), which is related to the highest mAP for these samples (as explained in Sec.~\ref{subsubsec:detection_precission}).
\begin{table*}[!b]
  \centering
  \caption{\label{tab:mape_bright} MAE, cMAE, and sMAPE calculated for the each class of microbes, and altogether, for Faster R-CNN and Cascade R-CNN with four different backbones on the \textit{higher-resolution} subset. cMAE metrics for each model are written in brackets.}
    \footnotesize
    \begin{tabular}{|lcccccccc|}
    \multicolumn{9}{c}{Microbe couting metrics for various models @ \textit{higher-resolution} subset} \\
    \multicolumn{1}{r}{} & \multicolumn{8}{c}{Faster R-CNN} \\
    \multicolumn{1}{l}{\multirow{2}[0]{*}{microbe}} & \multicolumn{2}{c}{\cellcolor[rgb]{ .851,  .851,  .851}ResNet-50} & \multicolumn{2}{c}{\cellcolor[rgb]{ .949,  .949,  .949}ResNet-101} & \multicolumn{2}{c}{\cellcolor[rgb]{ .851,  .851,  .851}ResNeXt-101} & \multicolumn{2}{c}{\cellcolor[rgb]{ .949,  .949,  .949}HRNet} \\
    \multicolumn{1}{l}{} & MAE   & sMAPE & MAE   & sMAPE & MAE   & sMAPE & MAE   & \multicolumn{1}{c}{sMAPE} \\
    \multicolumn{1}{l}{\textit{S. aureus}} & 0.42 & \cellcolor[rgb]{ .973,  .412,  .42}2.78\% & 0.66  & \cellcolor[rgb]{ .914,  .894,  .51}2.43\% & 0.39  & \cellcolor[rgb]{ .651,  .82,  .494}2.35\% & 0.67  & \multicolumn{1}{c}{\cellcolor[rgb]{ .992,  .776,  .49}2.55\%} \\
    \multicolumn{1}{l}{\textit{B. subtilis}} & 0.66  & \cellcolor[rgb]{ .984,  .612,  .459}1.96\% & 0.74  & \cellcolor[rgb]{ .992,  .753,  .486}1.88\% & 0.57  & \cellcolor[rgb]{ .973,  .412,  .42}2.07\% & 0.67  & \multicolumn{1}{c}{\cellcolor[rgb]{ .886,  .886,  .51}1.69\%} \\
    \multicolumn{1}{l}{\textit{P. aeruginosa}} & 1.21  & \cellcolor[rgb]{ .973,  .412,  .42}5.53\% & 1.35  & \cellcolor[rgb]{ 1,  .91,  .518}4.64\% & 1.50  & \cellcolor[rgb]{ .984,  .576,  .451}5.24\% & 1.28  & \multicolumn{1}{c}{\cellcolor[rgb]{ .506,  .776,  .486}4.37\%} \\
    \multicolumn{1}{l}{\textit{E. coli}} & 1.62  & \cellcolor[rgb]{ .973,  .412,  .42}5.40\% & 1.77  & \cellcolor[rgb]{ .988,  .663,  .471}4.83\% & 1.79  & \cellcolor[rgb]{ .976,  .914,  .514}4.18\% & 1.63  & \multicolumn{1}{c}{\cellcolor[rgb]{ 1,  .906,  .518}4.27\%} \\
    \multicolumn{1}{l}{\textit{C. albicans}} & 0.93  & \cellcolor[rgb]{ .973,  .412,  .42}2.98\% & 0.56  & \cellcolor[rgb]{ .855,  .878,  .506}2.34\% & 0.60  & \cellcolor[rgb]{ .996,  .784,  .494}2.58\% & 0.53  & \multicolumn{1}{c}{\cellcolor[rgb]{ .675,  .827,  .498}2.23\%} \\
    \midrule
    altogether & 4.75 (4.84)  & \cellcolor[rgb]{ .988,  .671,  .471}5.32\% & 4.44 (5.08)  & \cellcolor[rgb]{ .973,  .412,  .42}5.54\% & 4.47 (4.85)   & \cellcolor[rgb]{ 1,  .882,  .51}5.14\% & 4.35 (4.78)  & \cellcolor[rgb]{ .537,  .788,  .49}4.92\% \\
    \midrule
    \multicolumn{1}{r}{} & \multicolumn{8}{c}{Cascade R-CNN} \\
    \multicolumn{1}{l}{\multirow{2}[0]{*}{microbe}} & \multicolumn{2}{c}{\cellcolor[rgb]{ .851,  .851,  .851}ResNet-50} & \multicolumn{2}{c}{\cellcolor[rgb]{ .949,  .949,  .949}ResNet-101} & \multicolumn{2}{c}{\cellcolor[rgb]{ .851,  .851,  .851}ResNeXt-101} & \multicolumn{2}{c}{\cellcolor[rgb]{ .949,  .949,  .949}HRNet} \\
    \multicolumn{1}{l}{} & MAE   & sMAPE & MAE   & sMAPE & MAE   & sMAPE & MAE   & \multicolumn{1}{c}{sMAPE} \\
    \multicolumn{1}{l}{\textit{S. aureus}} & 0.67  & \cellcolor[rgb]{ .973,  .412,  .42}2.51\% & 0.65  & \cellcolor[rgb]{ .855,  .878,  .506}2.40\% & 0.35  & \cellcolor[rgb]{ .984,  .631,  .463}2.48\% & 0.68  & \multicolumn{1}{c}{\cellcolor[rgb]{ .388,  .745,  .482}2.27\%} \\
    \multicolumn{1}{l}{\textit{B. subtilis}} & 0.78  & \cellcolor[rgb]{ .616,  .808,  .494}1.36\% & 0.71  & \cellcolor[rgb]{ .388,  .745,  .482}1.26\% & 0.45  & \cellcolor[rgb]{ .992,  .733,  .482}1.69\% & 0.68  & \multicolumn{1}{c}{\cellcolor[rgb]{ .973,  .412,  .42}1.97\%} \\
    \multicolumn{1}{l}{\textit{P. aeruginosa}} & 1.39  & \cellcolor[rgb]{ .973,  .412,  .42}5.04\% & 1.30  & \cellcolor[rgb]{ 1,  .867,  .51}4.59\% & 1.69  & \cellcolor[rgb]{ .388,  .745,  .482}4.31\% & 1.26  & \multicolumn{1}{c}{\cellcolor[rgb]{ .847,  .878,  .506}4.48\%} \\
    \multicolumn{1}{l}{\textit{E. coli}} & 1.71  & \cellcolor[rgb]{ .973,  .412,  .42}4.40\% & 1.69  & \cellcolor[rgb]{ .996,  .796,  .494}3.81\% & 1.84  & \cellcolor[rgb]{ .792,  .859,  .502}3.42\% & 1.58  & \multicolumn{1}{c}{\cellcolor[rgb]{ .388,  .745,  .482}3.03\%} \\
    \multicolumn{1}{l}{\textit{C. albicans}} & 0.51  & \cellcolor[rgb]{ .988,  .682,  .475}2.52\% & 0.50  & \cellcolor[rgb]{ .525,  .784,  .49}2.11\% & 0.35  & \cellcolor[rgb]{ .973,  .412,  .42}2.75\% & 0.49  & \multicolumn{1}{c}{\cellcolor[rgb]{ .388,  .745,  .482}2.05\%} \\
    \midrule
    altogether & 4.67 (5.06)  & \cellcolor[rgb]{ .973,  .412,  .42}5.15\% & 4.41 (4.85)  & \cellcolor[rgb]{ .992,  .753,  .486}5.07\% & 4.40 (4.68)  & \cellcolor[rgb]{ .855,  .878,  .506}4.99\% & 4.31 (4.69)  & \cellcolor[rgb]{ .388,  .745,  .482}\textbf{4.86\%} \\
    \bottomrule
    \end{tabular}
\end{table*}

\begin{table*}[!htb]
  \centering
  \caption{\label{tab:mape_lower-resolution} Same as in Table~\ref{tab:mape_bright} but for the \textit{lower-resolution} subset.}
    \footnotesize
    \begin{tabular}{|lcccccccc|}
    \multicolumn{9}{c}{Microbe couting metrics for various models @ \textit{lower-resolution} subset} \\
    \multicolumn{1}{r}{} & \multicolumn{8}{c}{Faster R-CNN} \\
    \multicolumn{1}{l}{\multirow{2}[0]{*}{microbe}} & \multicolumn{2}{c}{\cellcolor[rgb]{ .851,  .851,  .851}ResNet-50} & \multicolumn{2}{c}{\cellcolor[rgb]{ .949,  .949,  .949}ResNet-101} & \multicolumn{2}{c}{\cellcolor[rgb]{ .851,  .851,  .851}ResNeXt-101} & \multicolumn{2}{c}{\cellcolor[rgb]{ .949,  .949,  .949}HRNet} \\
    \multicolumn{1}{c}{} & MAE   & sMAPE & \multicolumn{1}{c}{MAE} & sMAPE & MAE   & sMAPE & MAE   & \multicolumn{1}{c}{sMAPE} \\
    \multicolumn{1}{l}{\textit{S. aureus}} & 0.54  & \cellcolor[rgb]{ .992,  .761,  .49}3.55\% & \multicolumn{1}{c}{0.37} & \cellcolor[rgb]{ .992,  .737,  .482}3.58\% & 0.36  & \cellcolor[rgb]{ .584,  .8,  .49}2.58\% & 0.27  & \multicolumn{1}{c}{\cellcolor[rgb]{ .463,  .765,  .486}2.35\%} \\
    \multicolumn{1}{l}{\textit{B. subtilis}} & 0.23  & \cellcolor[rgb]{ .996,  .792,  .494}0.63\% & \multicolumn{1}{c}{0.25} & \cellcolor[rgb]{ .875,  .886,  .51}0.54\% & 0.37  & \cellcolor[rgb]{ .973,  .412,  .42}0.79\% & 0.16  & \multicolumn{1}{c}{\cellcolor[rgb]{ .388,  .745,  .482}0.40\%} \\
    \multicolumn{1}{l}{\textit{P. aeruginosa}} & 0.55  & \cellcolor[rgb]{ .996,  .8,  .498}5.60\% & \multicolumn{1}{c}{0.39} & \cellcolor[rgb]{ .522,  .78,  .486}4.36\% & 0.46  & \cellcolor[rgb]{ .745,  .847,  .502}4.74\% & 0.39  & \multicolumn{1}{c}{\cellcolor[rgb]{ .467,  .769,  .486}4.27\%} \\
    \multicolumn{1}{l}{\textit{E. coli}} & 0.75  & \cellcolor[rgb]{ .992,  .71,  .478}4.71\% & \multicolumn{1}{c}{0.65} & \cellcolor[rgb]{ .808,  .867,  .506}3.82\% & 0.57  & \cellcolor[rgb]{ .698,  .835,  .498}3.62\% & 0.59  & \multicolumn{1}{c}{\cellcolor[rgb]{ .408,  .749,  .482}3.09\%} \\
    \multicolumn{1}{l}{\textit{C. albicans}} & 0.37  & \cellcolor[rgb]{ .996,  .8,  .494}1.49\% & \multicolumn{1}{c}{0.33} & \cellcolor[rgb]{ .388,  .745,  .482}1.08\% & 0.25  & \cellcolor[rgb]{ .835,  .875,  .506}1.22\% & 0.22  & \multicolumn{1}{c}{\cellcolor[rgb]{ .612,  .808,  .494}1.15\%} \\
    \midrule
    altogether & 1.83 (2.24)  & \cellcolor[rgb]{ .973,  .412,  .42}4.68\% & \multicolumn{1}{c}{1.69 (1.99)} & \cellcolor[rgb]{ .984,  .561,  .451}4.55\% & 1.92 (2.01)  & \cellcolor[rgb]{ .969,  .91,  .514}4.21\% & 1.54 (1.63)  & \cellcolor[rgb]{ .839,  .875,  .506}4.12\% \\
    \midrule
    \multicolumn{1}{r}{} & \multicolumn{8}{c}{Cascade R-CNN} \\
    \multicolumn{1}{l}{\multirow{2}[0]{*}{microbe }} & \multicolumn{2}{c}{\cellcolor[rgb]{ .851,  .851,  .851}ResNet-50} & \multicolumn{2}{c}{\cellcolor[rgb]{ .949,  .949,  .949}ResNet-101} & \multicolumn{2}{c}{\cellcolor[rgb]{ .851,  .851,  .851}ResNeXt-101} & \multicolumn{2}{c}{\cellcolor[rgb]{ .949,  .949,  .949}HRNet} \\
    \multicolumn{1}{l}{} & MAE   & sMAPE & \multicolumn{1}{c}{MAE} & sMAPE & MAE   & sMAPE & MAE   & \multicolumn{1}{c}{sMAPE} \\
    \multicolumn{1}{l}{\textit{S. aureus}} & 0.33  & \cellcolor[rgb]{ 1,  .851,  .506}3.45\% & 0.33  & \cellcolor[rgb]{ .973,  .412,  .42}3.94\% & 0.30  & \cellcolor[rgb]{ .957,  .906,  .514}3.29\% & 0.32  & \multicolumn{1}{c}{\cellcolor[rgb]{ .388,  .745,  .482}2.20\%} \\
    \multicolumn{1}{l}{\textit{B. subtilis}} & 0.35  & \cellcolor[rgb]{ .973,  .412,  .42}0.60\% & 0.31  & \cellcolor[rgb]{ .961,  .91,  .514}0.57\% & 0.32  & \cellcolor[rgb]{ .996,  .824,  .502}0.58\% & 0.31  & \multicolumn{1}{c}{\cellcolor[rgb]{ .388,  .745,  .482}0.49\%} \\
    \multicolumn{1}{l}{\textit{P. aeruginosa}} & 0.43  & \cellcolor[rgb]{ .992,  .725,  .482}6.57\% & 0.42  & \cellcolor[rgb]{ .973,  .412,  .42}6.97\% & 0.43  & \cellcolor[rgb]{ .925,  .898,  .51}6.06\% & 0.40  & \multicolumn{1}{c}{\cellcolor[rgb]{ .388,  .745,  .482}4.13\%} \\
    \multicolumn{1}{l}{\textit{E. coli}} & 0.60  & \cellcolor[rgb]{ .973,  .412,  .42}5.46\% & 0.58  & \cellcolor[rgb]{ .851,  .878,  .506}4.51\% & 0.57  & \cellcolor[rgb]{ .976,  .443,  .427}5.43\% & 0.52  & \multicolumn{1}{c}{\cellcolor[rgb]{ .388,  .745,  .482}3.05\%} \\
    \multicolumn{1}{l}{\textit{C. albicans}} & 0.25  & \cellcolor[rgb]{ .996,  .839,  .502}1.56\% & 0.24  & \cellcolor[rgb]{ .973,  .412,  .42}2.17\% & 0.22  & \cellcolor[rgb]{ .388,  .745,  .482}1.18\% & 0.21  & \multicolumn{1}{c}{\cellcolor[rgb]{ .718,  .839,  .498}1.32\%} \\
    \midrule
    altogether & 1.70 (1.96) & \cellcolor[rgb]{ .922,  .898,  .51}4.15\% & 1.62 (1.88)  & \cellcolor[rgb]{ .988,  .69,  .475}4.25\% & 1.60 (1.84)  & \cellcolor[rgb]{ .973,  .412,  .42}4.31\% & 1.57 (1.76)  & \cellcolor[rgb]{ .388,  .745,  .482}\textbf{3.81\%} \\
    \bottomrule
    \end{tabular}
\end{table*}

\subsection{Anchor tuning}
\label{subsubsec:hyperparameters_tuning}

Hyperparameters are the variables which determines the network structure itself (e.g the number of hidden layers) or its training process (e.g. the learning rate or the number of epochs). The models tested in this study are already provided with the default (pre-optimized) learning hyperparameters values\cite{bib:mmdetection2019} and network architectures, which is specific for each model and backbone used. However, it may be interesting to verify if proper tuning of some parameters, to adjust  them to our specific dataset, gives us some better results in terms of lower counting error metrics. 

One of the crucial parameter that impact object detection performance is the anchor boxes setup. Anchors are some initial guessed bounding boxes that are assumed at the start of the detection procedure. The proper selection of anchors is a broad topic, currently investigated by many researchers. For example, automatic anchor selection techniques, i.e. which learn meta-anchors during network training, have been proposed recently\cite{bib:anchors_learning1,bib:anchors_learning2}.

Here, basic approach is applied by simply variating the anchors' parameters and training the Cascade R-CNN with HRNet model on \textit{higher-resolution} subset for each setup separately. In Table~\ref{tab:mape_hyperparams} there are presented results (counting errors) for the various anchors' parameters variations: (1) the aspect ratio (i.e. $x$ vs $y$ size), (2) the list of scales, and (3) the stride modification. While the anchors setup has an impact on the detection performance, it is hard to significantly overcome results for default settings. This suggests that chosen pre-tuned models are well adjusted to work on the AGAR dataset. The only significant impact comes from modifying anchors' strides, causing noteworthy decrease of MAE with slightly higher sMAPE. This means that the tuning of the stride improves the results for high populated samples. Moreover, narrowing the anchors' ratio range from $[0.5, 2.0]$ to $[0.75, 1.5]$ improves sMAPE from 4.86\% to 4.68\%. It is expected as a typical bounding box for microbe colony has a square-like shape.

\begin{table*}[!bt]
  \centering
  \caption{\label{tab:mape_hyperparams} Counting error metrics for various setups of anchors hyperparameters. Narrowing the range of anchors ratio lowers the sMAPE, while modifying anchors stride significantly improves MAE.}
  \footnotesize
    \begin{tabular}{l|cc|cc|cc|cc}
    \multicolumn{9}{c}{Model (Cascade(HRNet) @ \textit{higher-resolution} subset) hyperparameters optimization} \\
    \rowcolor[rgb]{ .949,  .949,  .949} \multicolumn{1}{p{6em}|}{parameter
    \newline{} modification} & \multicolumn{2}{p{7.5em}|}{without modification (reference)} & \multicolumn{2}{p{7em}|}{anchor ratio \newline{}variation } & \multicolumn{2}{p{7em}|}{anchor scaling variation} & \multicolumn{2}{p{7em}}{anchor stride list tuning} \\
    \multirow{2}[0]{*}{metrics} & MAE   & sMAPE & MAE   & sMAPE & MAE   & sMAPE & MAE   & sMAPE \\
          & \cellcolor[rgb]{ .933,  .902,  .514}4.31 & \cellcolor[rgb]{ .988,  .643,  .467}4.86\% & \cellcolor[rgb]{ .925,  .898,  .51}4.28 & \cellcolor[rgb]{ .996,  .788,  .494}4.68\% & \cellcolor[rgb]{ .949,  .906,  .514}4.35 & \cellcolor[rgb]{ .984,  .592,  .455}4.92\% & \cellcolor[rgb]{ .388,  .745,  .482}2.52 & \cellcolor[rgb]{ .973,  .412,  .42}5.14\% \\
    \end{tabular}%
  \label{tab:addlabel}
\end{table*}%

\subsection{\textit{Empty} and \textit{uncountable} plates}
\label{subsubsec:empty_uncountable}

The performance of trained Cascade R-CNN with HRNet backbone was evaluated additionally on the \textit{empty} and \textit{uncountable} samples of the AGAR dataset, for both major background categories: \textit{lower}- and \textit{higher-resolution} (without the \textit{vague} subset). In this case only qualitative results are provided, because ground truth annotations are available only for \textit{countable} plates.

\begin{figure}[!hbt]
\centering
\includegraphics[width=.7\textwidth]{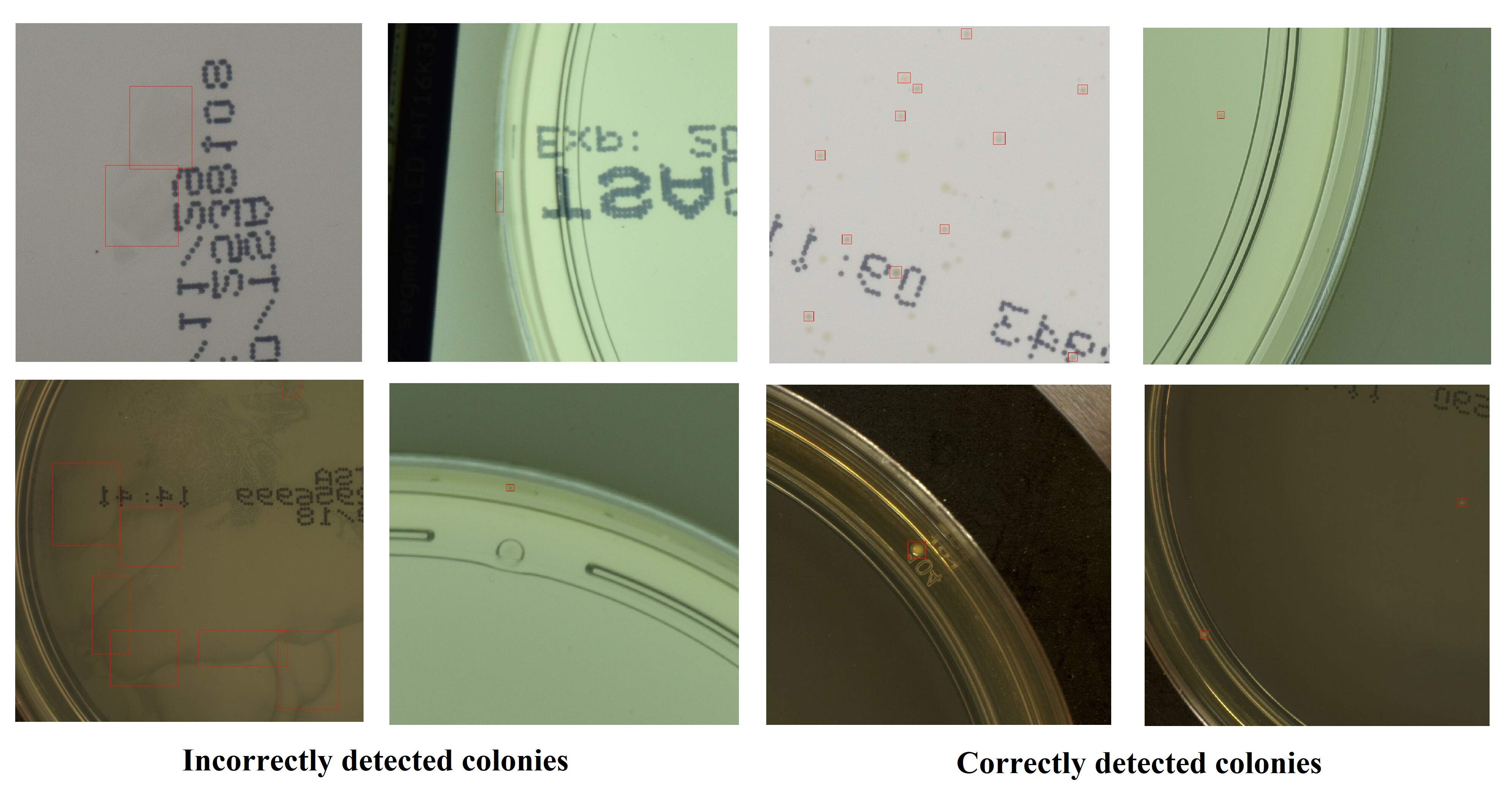}
\caption{\label{fig:empty} Examples of predictions for a few \textit{empty} samples.}
\end{figure}

Testing \textit{empty} samples allows us to investigate the false positive predictions (the model finds colonies on plates without microbes grown). The most common FPs are: discoloration and substrate loss, plate edge stains, formed blisters and water droplets. Also note that in some situations correct colonies are found, but they are hardly visible, so not marked by annotators. A few examples of such cases are presented in Fig.~\ref{fig:empty}.

\begin{figure}[!htb]
\centering
\includegraphics[width=.9\textwidth]{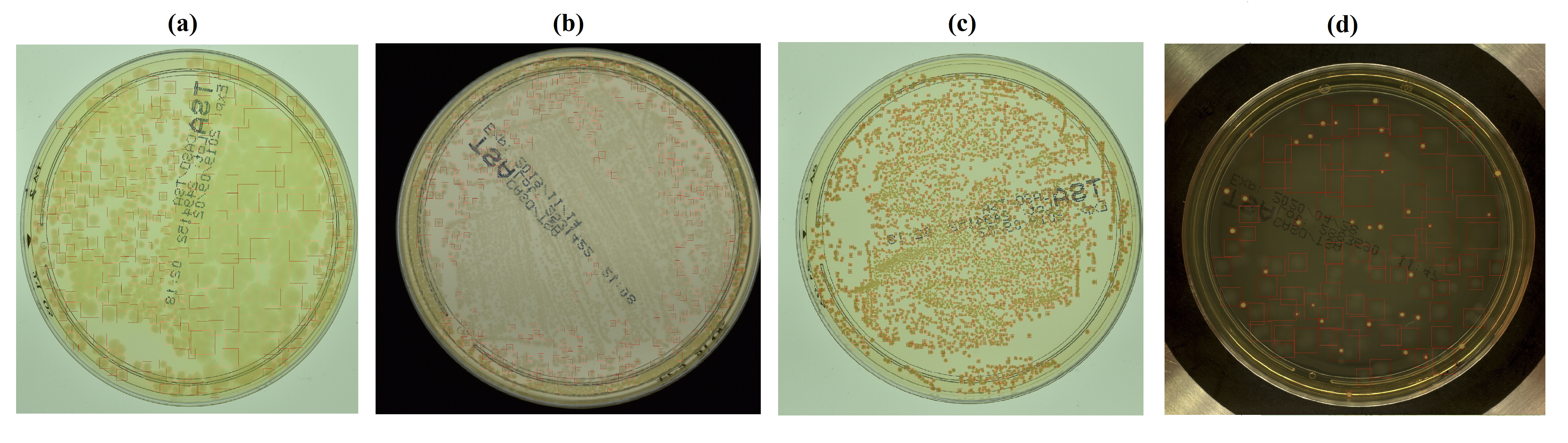}
\caption{\label{fig:nc} Examples of predictions for \textit{uncountable} cases.}
\end{figure}

On the other hand, tests performed on \textit{uncountable} samples enable us to check how the detector (trained on annotated data with max. 300 colonies) performs with very crowded plates. Here, the obvious problem for the detector is the so-called microbial lawn -- meaning individual colonies fused together and spread evenly over the agar surface (Fig.~\ref{fig:nc}(a)). Another similar challenge is recognizing (and consequently counting) individual colonies in  growth bacterial streak, as presented in Fig.~\ref{fig:nc}(b). Very small colonies (of a size of few pixels) are difficult to detect, however, due to their large separation, the model shows a great potential to roughly estimate the number of colonies even for very crowded plates -- at best, it detects 2782 colonies (Fig.~\ref{fig:nc}(c)). The last example (Fig.~\ref{fig:nc}(d)) presents the sample with two different species grown on one agar plate. The model is able to detect even hardly visible colonies, although it also recognizes some discolorations as a microbe.

\begin{figure}[!ht]
\centering
\includegraphics[width=.65\textwidth]{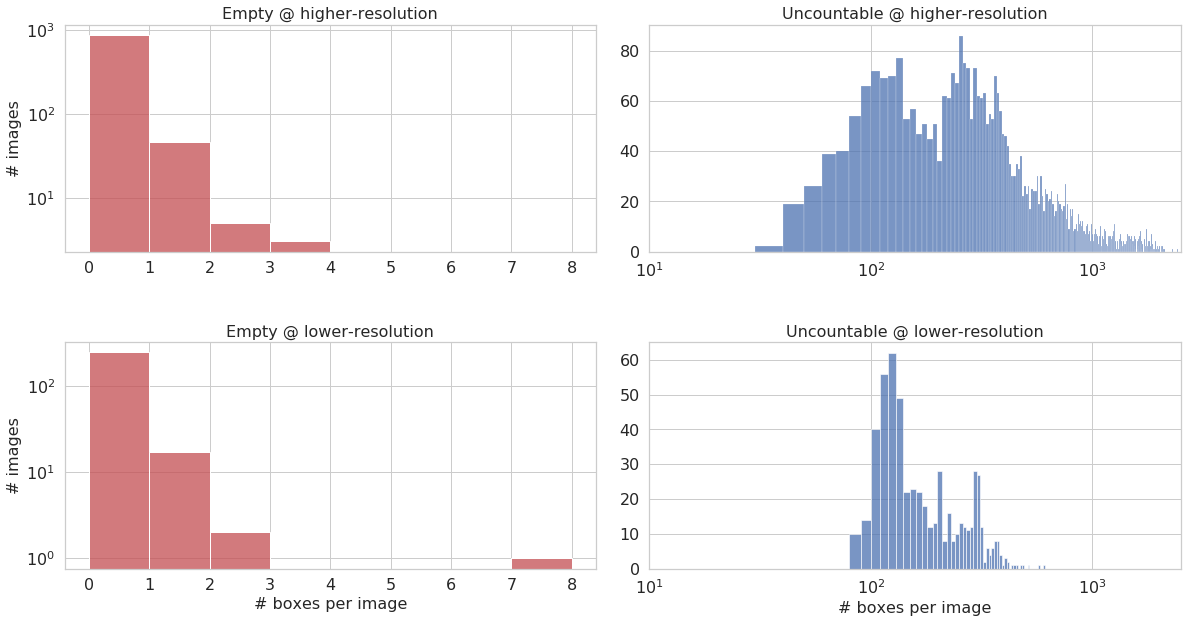}
\caption{\label{fig:hist_nc_empty} Distribution of number of predicted colonies for \textit{higher}- and \textit{lower-resolution} photos in both \textit{empty} and \textit{uncountable} edge cases.}
\end{figure}

The model performs quite well for detecting colonies in both edge cases. However, it handles \textit{empty} plates slightly better. The false positive predictions occur only for about 6-7\% of all \textit{empty} dishes. Note, however, that in some cases there are actually some formed colonies missed by the annotators (like in the examples in Fig.~\ref{fig:empty}). In the case of \textit{uncountable} subset, the number of colonies are very often underestimated. The model predicts less than 300 colonies for 41\% and 85\% samples for \textit{higher}- and \textit{lower-resolution}, respectively. However, only 7\% and 4\% are estimated to have less than 100 colonies. Note that it was harder to annotate the \textit{lower-resolution} images (because of low contrast), so some samples labeled as \textit{uncountable} may be actually within the defined countable range. Moreover, as discussed before, the sMAPE metric used for threshold optimization causes the model to be more accurate for samples with smaller number of colonies on a plate. Detailed distributions of counted colonies can be found in Fig.~\ref{fig:hist_nc_empty}.

\subsection{Double thresholding approach}
\label{subsubsec:combined_model}

Looking at the obtained results for microbe counting (see Fig.~\ref{fig:gt_vs_pred_4_types}), one can conclude that the trained models underestimate the number of colonies for more numerous samples. This is due to the distribution of our data, presented in Fig.~\ref{fig:count_histograms}, with more than 84\% of samples consisting of photos of plates with less than 50 grown colonies. To remedy this, a double thresholding approach was investigated. Two sets of the probability and NMS thresholds are established based on training data: {\it general} (tuned to all samples) and {\it auxiliary} (tuned to samples with more than 50 colonies). The {\it mixed} approach uses {\it general} thresholds by default, however, if the model predicts more than 50 colonies, the {\it auxiliary} ones are applied for a given sample.

\begin{table}[!hb]
\centering
\footnotesize
\caption{ \label{tab:mape_combined} Counting error metrics for Cascade with HRNet model in various setups of thresholds scenario.}
\begin{tabular}{@{}|c|lll|@{}}
\toprule
\rowcolor[HTML]{EFEFEF} 
\textbf{Parameters}                                                                             & \multicolumn{1}{c|}{\cellcolor[HTML]{EFEFEF}\textbf{Thresholds settings}}                     & \multicolumn{1}{c|}{\cellcolor[HTML]{EFEFEF}\textbf{MAE}} 
& \multicolumn{1}{c|}{\cellcolor[HTML]{EFEFEF}\textbf{sMAPE}} 
\\ \midrule
\cellcolor[HTML]{EFEFEF}                                                                        & \begin{tabular}[c]{@{}l@{}} general \end{tabular}                                 & 4.31 & \cellcolor[rgb]{ 1,  .922,  .518} 4.86\% \\
\cellcolor[HTML]{EFEFEF}                                                                        & \begin{tabular}[c]{@{}l@{}} auxiliary \end{tabular}                             
& 3.49 & \cellcolor[rgb]{ .973,  .412,  .42} 7.64\%\\
\multirow{-3}{*}{\cellcolor[HTML]{EFEFEF}\textbf{\begin{tabular}[c]{@{}c@{}}Cascade HRNet\\ \textit{higher-resolution}\end{tabular}}} & \begin{tabular}[c]{@{}l@{}}mixed\end{tabular} 
& 2.35 &  \cellcolor[rgb]{ .388,  .745,  .482} 4.06\%
\\ \midrule
\cellcolor[HTML]{EFEFEF}                                                                        & \begin{tabular}[c]{@{}l@{}} general \end{tabular} & 1.57
& \cellcolor[rgb]{ 1,  .922,  .518} 3.81\%\\
\cellcolor[HTML]{EFEFEF}                                                                        & \begin{tabular}[c]{@{}l@{}} auxiliary \end{tabular}  
& 1.50    & \cellcolor[rgb]{ .973,  .412,  .42} 4.97\% \\
\multirow{-3}{*}{\cellcolor[HTML]{EFEFEF}\textbf{\begin{tabular}[c]{@{}c@{}}Cascade HRNet\\ \textit{lower-resolution}\end{tabular}}}  & \begin{tabular}[c]{@{}l@{}}mixed\end{tabular} 
& 1.49 &\cellcolor[rgb]{ .388,  .745,  .482} 3.67\%                                   \\ \bottomrule
\end{tabular}
\end{table}

\begin{figure}[!h]
\centering
\includegraphics[width=.65\textwidth]{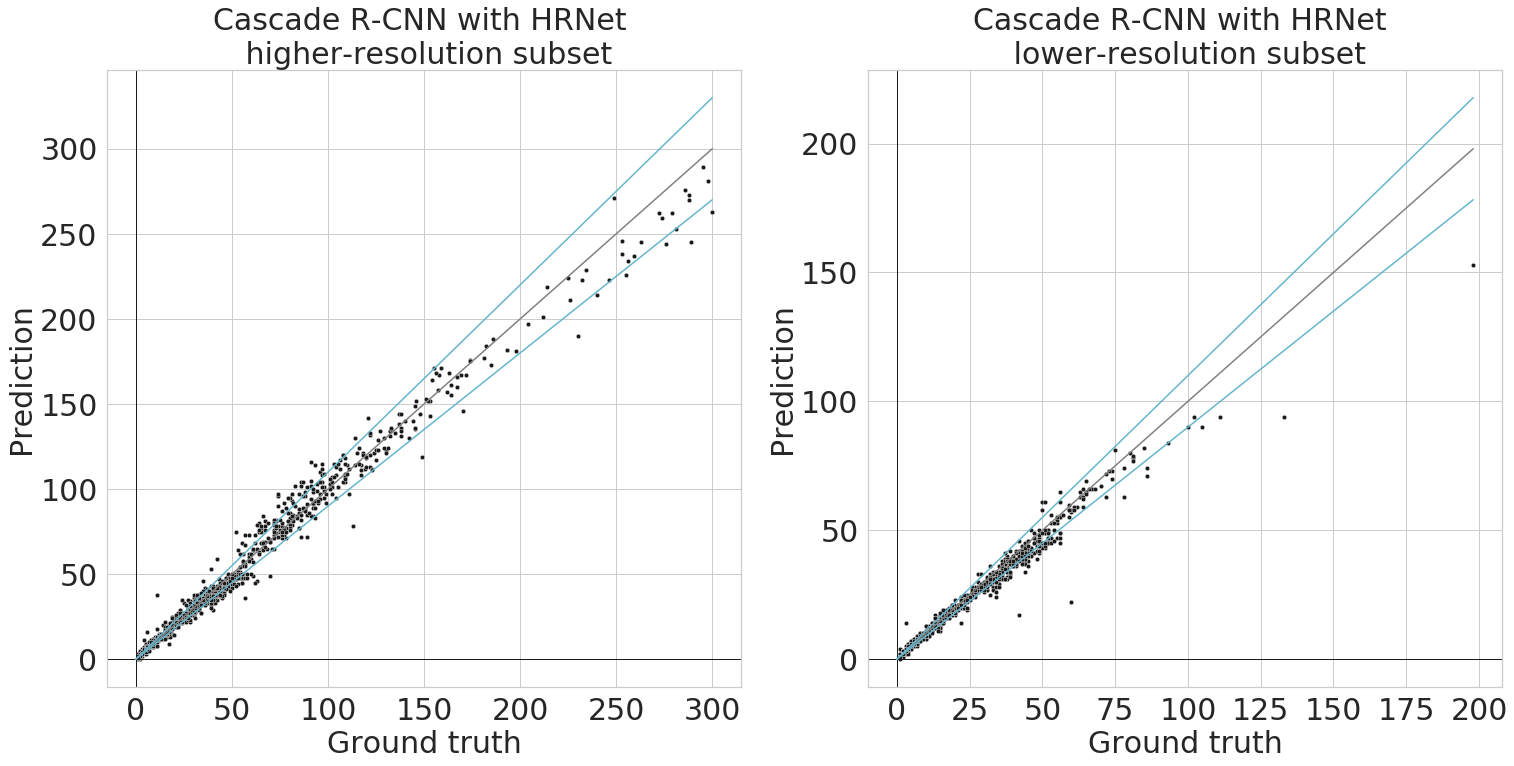}
\caption{\label{fig:gt_vs_pred_threshold_combined}Performance of microbial colony counting on two images subgroups using Cascade R-CNN with HRNet for double thresholding approach.}
\end{figure}

The results for different thresholds settings are presented in Table~\ref{tab:mape_combined}. As one can expect, {\it auxiliary} thresholds taken individually increase sMAPE for test data, but they also decrease MAE. However, {\it mixed} approach gives a good balance, and works slightly better than single {\it general} thresholding for all samples in the AGAR dataset. Its performance for microbe counting is shown in Fig.~\ref{fig:gt_vs_pred_threshold_combined}. 

\newpage

\subsection{Transfer learning}
\label{subsubsec:transfer_tearning}
Transfer learning is a technique used to improve learning a new task by transferring the knowledge from related problems\cite{bib:torrey2010transfer, bib:Pan2010ASO}. It can be used to adjust the model trained on the AGAR dataset for new type of samples (different microbes, lighting condition etc.), unseen by the detector, if the new data is too small to perform efficient training solely on it. Here, the ability of a pre-trained model to learn new features from photos taken in different data acquisition setup is checked.

A transfer learning experiment is proposed by using \textit{vague} (the smallest and the most difficult) subset to mimic unseen data, and assigning (much larger) \textit{bright+dark} subsets as a primary dataset used to initially train the model. In the first approach, the Cascade R-CNN with HRNet model pre-trained on the primary data is taken and then additional learning of the whole network using solely the \textit{vague} samples is performed. This leads to a new model, which (just after a few epochs) perfectly recognizes microbes on the \textit{vague} subset. However, it loses the performance on the primary dataset. 

To overcome this, i.e. make a model to learn new features but keep its ability to recognize the primary ones, one can update only a few last layers of the neural network, keeping the rest frozen~\cite{bib:yosinski2014transferable}. The results of this technique is presented in Table~\ref{tab:mape_fine-tunning}. With the increasing number of unfrozen layers, the counting metrics (MAE and sMAPE) significantly improve for the \textit{vague} subset, but get slightly worse for the primary data. Unfreezing more than 128 layers does not further improve the performance on the \textit{vague} photos, but makes the model less capable of recognizing the samples from \textit{bright} and \textit{dark} subsets.

\begin{table*}[!h]
  \centering
  \caption{\label{tab:mape_fine-tunning} Transfer learning for the model trained on the \textit{bright+dark} subsets to detect microbes on the \textit{vague} one, by updating only part of last layers  i.e. all other weights are frozen. We observe that with increasing the number of unfrozen layers, the counting metrics for the \textit{vague} subset improves, while for the \textit{bright+dark} one---decreases.}
    \footnotesize
    \begin{tabular}{lrrrrrrrrrr}
    \multicolumn{11}{c}{Fine-tuning of Cascade R-CNN (HRNet) model trained on \textit{bright+dark} subset to capture \textit{vague} subset} \\
    \rowcolor[rgb]{ .851,  .851,  .851} \# unfrozen layers & \multicolumn{2}{c}{\cellcolor[rgb]{ .949,  .949,  .949}0 (reference)} & \multicolumn{2}{c}{16} & \multicolumn{2}{c}{\cellcolor[rgb]{ .949,  .949,  .949}32} & \multicolumn{2}{c}{64} & \multicolumn{2}{c}{\cellcolor[rgb]{ .949,  .949,  .949}128} \\
    metrics for: & \multicolumn{1}{c}{MAE} & \multicolumn{1}{c}{sMAPE} & \multicolumn{1}{c}{MAE} & \multicolumn{1}{c}{sMAPE} & \multicolumn{1}{c}{MAE} & \multicolumn{1}{c}{sMAPE} & \multicolumn{1}{c}{MAE} & \multicolumn{1}{c}{sMAPE} & \multicolumn{1}{c}{MAE} & \multicolumn{1}{c}{sMAPE} \\
    \textit{bright+dark} subset & 4.31  & \cellcolor[rgb]{ .388,  .745,  .482}4.86\% & 3.92  & \cellcolor[rgb]{ .424,  .753,  .482}5.19\% & 7.36  & \cellcolor[rgb]{ 1,  .918,  .518}10.73\% & 7.39  & \cellcolor[rgb]{ 1,  .914,  .518}10.77\% & 7.21  & \cellcolor[rgb]{ 1,  .906,  .518}11.38\% \\
    \textit{vague} subset & 15.27 & \cellcolor[rgb]{ .973,  .412,  .42}39.55\% & 14.79 & \cellcolor[rgb]{ .976,  .431,  .424}38.49\% & 4.12  & \cellcolor[rgb]{ .949,  .906,  .514}9.84\% & 3.32  & \cellcolor[rgb]{ .808,  .863,  .506}8.59\% & 2.51 & \cellcolor[rgb]{ .439,  .757,  .482}5.34\% \\
    \end{tabular}%
\end{table*}%

\subsection{Training on the whole dataset at once}
\label{subsubsec:all_training}
Above, the results of transfer-learned version of Cascade R-CNN with HRNet on the AGAR dataset are reported. 
The conducted studies showed that the use of such transfer-learning methods to overcome some deficiencies introduced to our dataset (e.g. by removing the \textit{vague} part) might be not sufficient to prepare a model capable for general predictions. 
To improve the model's ability to generalize, the model was trained on the entire dataset (combined training subsets of \textit{lower}- and \textit{higher-resolution} images). Metrics evaluated for this case (see Table~\ref{tab:mape-all}) show that both Cascade R-CNN and Faster R-CNN models have great ability to capture different datasets \textit{at once}. This was verified by testing such generalized model on test data for each subset separately, as presented in Table~\ref{tab:mape-all}.

Let us compare the best result for the transfer technique presented in Subsection~\ref{subsubsec:transfer_tearning} (Table~\ref{tab:mape_fine-tunning}), i.e. sMAPE of 11.38\% (\textit{bright+dark}) and 5.34\% (\textit{vague}), with the results from the Table~\ref{tab:mape-all} for Cascade R-CNN (HRNet) giving about 4.98\% for the \textit{bright+dark}, and 1.96\% for the \textit{vague} subset. The models trained on all combined subsets at once performed better than these obtained using transfer of learning presented in the subsection \ref{subsubsec:transfer_tearning}. However, the transfer learning technique gives the possibility to quickly adopt a model for a new subset, without having access to the primary data the model was trained on.

\begin{table*}[!hb]
  \centering
  \caption{\label{tab:mape-all} Results for the Faster and Cascade R-CNN models with ResNet-50 and HRNet backbones trained on the whole dataset (\textit{higher}- and \textit{lower-resolution}), and tested on different subsets respectively. Evaluated metrics show that the both models have great ability to capture various datasets simulateously.}
  \footnotesize
    \begin{tabular}{lcccccccccccc}
    \multicolumn{13}{c}{Microbe couting metrics for various models trained on the all dataset} \\
    \multirow{2}[0]{*}{model} & \multicolumn{12}{c}{Faster R-CNN} \\
          & \multicolumn{6}{c}{ResNet-50}                 & \multicolumn{6}{c}{HRNet} \\
    \rowcolor[rgb]{ .851,  .851,  .851} tested on: & \multicolumn{2}{c}{\cellcolor[rgb]{ .949,  .949,  .949}\textit{bright+dark}} & \multicolumn{2}{c}{\textit{vague}} & \multicolumn{2}{c}{\cellcolor[rgb]{ .949,  .949,  .949}\textit{lower-resolution}} & \multicolumn{2}{c}{\textit{bright+dark}} & \multicolumn{2}{c}{\cellcolor[rgb]{ .949,  .949,  .949}\textit{vague}} & \multicolumn{2}{c}{\textit{lower-resolution}} \\
    \multirow{2}[1]{*}{metrics} & MAE   & sMAPE & MAE   & sMAPE & MAE   & sMAPE & MAE   & sMAPE & MAE   & sMAPE & MAE   & sMAPE \\
          & 4.87  & \cellcolor[rgb]{ .973,  .412,  .42}5.70\% & 1.10  & \cellcolor[rgb]{ .549,  .788,  .49}2.60\% & 1.96  & \cellcolor[rgb]{ .992,  .722,  .482}4.90\% & 4.25  & \cellcolor[rgb]{ .992,  .737,  .482}4.86\% & 0.88  & \cellcolor[rgb]{ .388,  .745,  .482}1.96\% & 1.54  & \cellcolor[rgb]{ .878,  .886,  .51}3.90\% \\
    \midrule
    \multirow{2}[1]{*}{model} & \multicolumn{12}{c}{Cascade R-CNN} \\
          & \multicolumn{6}{c}{ResNet-50}                 & \multicolumn{6}{c}{HRNet} \\
    \rowcolor[rgb]{ .851,  .851,  .851} tested on & \multicolumn{2}{c}{\cellcolor[rgb]{ .949,  .949,  .949}\textit{bright+dark}} & \multicolumn{2}{c}{\textit{vague}} & \multicolumn{2}{c}{\cellcolor[rgb]{ .949,  .949,  .949}\textit{lower-resolution}} & \multicolumn{2}{c}{\textit{bright+dark}} & \multicolumn{2}{c}{\cellcolor[rgb]{ .949,  .949,  .949}vague} & \multicolumn{2}{c}{\textit{lower-resolution}} \\
    \multirow{2}[0]{*}{metrics} & MAE   & sMAPE & MAE   & sMAPE & MAE   & sMAPE & MAE   & sMAPE & MAE   & sMAPE & MAE   & sMAPE \\
          & 4.97  & \cellcolor[rgb]{ .973,  .412,  .42}5.72\% & 1.33  & \cellcolor[rgb]{ .514,  .78,  .486}2.42\% & 1.71  & \cellcolor[rgb]{ .996,  .816,  .498}4.48\% & 4.29  & \cellcolor[rgb]{ .988,  .655,  .467}4.98\% & 0.83  & \cellcolor[rgb]{ .388,  .745,  .482}1.96\% & 1.66  & \cellcolor[rgb]{ .906,  .894,  .51}3.82\% \\
    \end{tabular}%
\end{table*}%

\bibliography{supplementary.bib}